\documentclass[twoside]{article}

\usepackage[accepted]{arxiv}

\usepackage{balance}
\usepackage[utf8]{inputenc} 
\usepackage[T1]{fontenc}    
\usepackage[round]{natbib}
\usepackage{hyperref}
\hypersetup{
colorlinks=true,
    citecolor = {blue}
}
\usepackage{url}            
\usepackage{booktabs}       
\usepackage{amsfonts}       
\usepackage{nicefrac}       
\usepackage{microtype}      

\usepackage{amsmath,mathrsfs,amssymb}
\usepackage{algpseudocode,algorithm,algorithmicx}
\usepackage{graphicx}
\usepackage{rotating}
\usepackage{wrapfig}

\DeclareMathOperator{\diag}{diag}
\DeclareMathOperator{\supp}{supp}

\DeclareMathOperator{\subspan}{span}

\DeclareMathOperator{\nnz}{nnz}

\newcommand{\T}{\rm T}
\newcommand{\wt}{\widetilde}
\newcommand{\wh}{\hat}

\newcommand{\R}{\mathbb{R}}
\newcommand{\I}{\mathcal{I}}
\newcommand{\J}{\mathcal{J}}
\newcommand{\bsmat}{\left[\begin{smallmatrix} }
\newcommand{\esmat}{\end{smallmatrix}\right] }

\newtheorem{example}{Example}
\newtheorem{theorem}{Theorem}
\newtheorem{lemma}{Lemma}

\newtheorem{remark}{Remark}

\begin{document}

%
\runningtitle{An Inverse-free Truncated Rayleigh-Ritz Method for Sparse Generalized Eigenvalue Problem}

%

\twocolumn[

\aistatstitle{\vspace{-0.1in}An Inverse-free Truncated Rayleigh-Ritz Method for \\ Sparse Generalized Eigenvalue Problem\vspace{-0.1in}}

\aistatsauthor{ Yunfeng Cai and Ping Li }

\aistatsaddress{  Cognitive Computing Lab\\
Baidu Research\\
No. 10 Xibeiwang East Road, Beijing 100085, China\\
10900 NE 8th St. Bellevue, WA 98004, USA\\ \{caiyunfeng, liping11\}@baidu.com}]

\begin{abstract}\vspace{-0.15in}
This paper considers the sparse generalized eigenvalue problem (SGEP), which aims to find the leading eigenvector with at most $k$ nonzero entries.
SGEP naturally arises in many applications in machine learning, statistics, and scientific computing, for example, the sparse principal component analysis (SPCA), the sparse discriminant analysis (SDA), and the sparse canonical correlation analysis (SCCA).
 In this paper, we focus on the development of a three-stage algorithm named {\em inverse-free truncated Rayleigh-Ritz method} ({\em IFTRR})
 to efficiently solve  SGEP.
In each iteration of IFTRR, only a small number of matrix-vector products is required.
This makes IFTRR well-suited for large scale problems.
Particularly, a new truncation strategy is proposed, which is able to find the support set of the leading eigenvector effectively.
Theoretical results are developed to explain why IFTRR works well.
Numerical simulations demonstrate the merits of IFTRR.
\end{abstract}

\vspace{-0.15in}
\section{Introduction}\label{sec:intro}
\vspace{-0.05in}

Given a matrix pair $(\wt{A},\wt{B})$,
where $\wt{A}$, $\wt{B}$ are both $p$-by-$p$ symmetric matrices and $\wt{B}$ is (semi) positive definite,
the sparse generalized eigenvalue problem (sparse GEP, or SGEP) aims to maximize the Rayleigh quotient $\frac{v^{\T}\wt{A}v}{v^{\T}\wt{B}v}$
with $v\in\R^p$ having no more than $k$ nonzero entries, where $k\ll p$.
Mathematically, SGEP can be formulated as the following optimization problem:
\begin{equation}\label{main}
\max_{v\in\R^{p}} \frac{v^{\T}\wt{A}v}{v^{\T}\wt{B}v},\quad \mbox{subject to}\quad \|v\|_0\le k,
\end{equation}
where
$\|v\|_0$ denotes the $\ell_0$-norm of $v$, which is the number of nonzero entries of $v$.
In many applications, such as sparse principle component analysis (SPCA)~\citep{zou2006sparse},
sparse discriminant analysis (SDA)~\citep{clemmensen2011sparse},
and sparse canonical correlation analysis (SCCA)~\citep{witten2009penalized} in high dimensional statistical analysis,
the matrices $\wt{A}$ and $\wt{B}$ usually can be decomposed as follows:
\begin{align}
\wt{A}=A+E,\qquad \wt{B}=B+F,
\end{align}
where ${A}$, ${B}$ are both symmetric, $B$ is positive definite, $E$, $F$ are symmetric perturbations due to finite sample~estimation.

Next, we consider the following two concrete examples of SGEP, which arise from Sparse Fisher's discriminant analysis (SFDA) and SCCA, respectively.
\begin{example}\label{eg:fda}
Given a data matrix $X\in\R^{n\times p}$ (i.e., $n$ observations with $p$ features),  each row belongs  to one of $K$ classes.
Denote $\mathcal{C}_k\subset \{1,2,\dots,n\}$ the indices of the observations in the $k$-th class, $n_k=|\mathcal{C}_k|$,
$\bar{x}_k=\sum_{i\in\mathcal{C}_k}\frac{X_{(i,:)}}{n_k}$.
Then we may use
\begin{align*}
\wt{\Sigma}_b&=\sum_{k=1}^K\frac{n_k\bar{x}_k^{\T}\bar{x}_k}{n},\\
\wt{\Sigma}_w&=\frac{1}{n}\sum_{k=1}^K\sum_{i\in\mathcal{C}_k}(X_{(i,:)}-\bar{x}_k)^{\T}(X_{(i,:)}-\bar{x}_k),
\end{align*}
as the estimators for the between class variance $\Sigma_b$ and the within class variance $\Sigma_w$, respectively.
SFDA intends to find a sparse leading discriminant vector $v$ that maximizes $\frac{v^{\T}\wt{\Sigma}_bv}{v^{\T}\wt{\Sigma}_wv}$,
which can be formulated as an SGEP with $\wt{A}=\wt{\Sigma}_b$, $\wt{B}=\wt{\Sigma}_w$.
\end{example}

\begin{example}\label{eg:cca}
Let $X\in\R^{p/2}$ and $Y\in\R^{p/2}$ be two random variables, $\Sigma_{xx}$, $\Sigma_{yy}$, $\Sigma_{xy}$ be the covariance matrices for $X$, $Y$, the cross-covariance matrix between $X$ and $Y$, respectively,
 ${\wh{\Sigma}}_{xx}$, $\wh{\Sigma}_{yy}$, $\wh{\Sigma}_{xy}$ be estimators for $\Sigma_{xx}$, $\Sigma_{yy}$, $\Sigma_{xy}$, respectively.
SCCA aims to maximize $v_x^{\T}\wh{\Sigma}_{xy}v_y$,
subject to $v_x^{\T}\wh{\Sigma}_{xx}v_x=1$, $v_y^{\T}\wh{\Sigma}_{yy}v_y=1$, $\|v_x\|_0\le s_x$, $\|v_y\|_0\le s_y$,
where $s_x$ and $s_y$ are two small integers.
Such a problem can be reformulated as an SGEP with $\wt{A}=\bsmat 0 & \wh{\Sigma}_{xy}\\ \wh{\Sigma}_{xy}^{\T} & 0\esmat$, $\wt{B}=\bsmat \wh{\Sigma}_{xx} & 0\\ 0 & \wh{\Sigma}_{yy}\esmat$,  $v=\bsmat v_x\\ v_y\esmat$.
\end{example}

\vspace{-0.02in}

\textbf{Computational challenges.}
The SGEP can be computationally challenging. Recall that in our model we have $\wt{A}=A+E$, $\wt{B}=B+F$. Denote the eigenvalues and the corresponding eigenvectors of $Av=\lambda Bv$
by $\lambda_1\ge \lambda_2\ge \dots\ge \lambda_p$ and
$v_1,v_2,\dots, v_p$, respectively.
We call $\lambda_1$ the leading eigenvalue of $(A,B)$, $v_1$ the leading eigenvector of $(A,B)$,
and $(\lambda_1,v_1)$ the leading eigenpair of $(A,B)$. In this paper, the leading eigenvector $v_1$ is assumed to be sparse,
i.e., $\|v_1\|_0\ll p$. Overall, the task of SGEP is essentially to find an approximation of $v_1$
via $(\wt{A},\wt{B})$, without knowing $E$, $F$.

Due to finite number of samples, the perturbations $E$ and $F$ may be large, and consequently, the leading eigenvector of $(\wt{A},\wt{B})$ may  not be a good approximation of the true $v_1$.  Furthermore, in a high dimensional setting, $\wt{A}$ and $\wt{B}$ can be both ill-conditioned (or singular), meaning that infinity eigenvalue $\infty$ and indeterminate eigenvalue $0/0$ occur.  In the presence of rounding-off error,
numerical algorithms may fail to detect the singularity due to ill-conditioning.
As a result, one may not be able to compute accurate or meaningful eigenvalues and eigenvectors. In fact, the optimization problem~\eqref{main} is essentially a subset selection problem,
which is known to be NP-hard~\citep{Proc:Moghaddam_NIPS05,Proc:Moghaddam_ICML06}.

Here, it is  worth mentioning that,  when $B=\wt{B}=I_p$, the problem~\eqref{main} is reduced to the so-called {\em sparse eigenvalue problem} (SEP), also known as {\em sparse principal component analysis} (SPCA). Obviously SGEP can be substantially more challenging than SEP (SPCA).

\textbf{Related work.}
A variety of numerical methods have been proposed for SEP. 
Existing algorithms of SEP are mostly optimization approaches, which are based on relaxation, or penalization, or both.
The $\ell_1$-norm relaxation, inspired by LASSO, is first studied in~\citep{jolliffe2003modified} and called SCoTLASS.
In~\citep{witten2009penalized}, a penalized matrix decomposition method is proposed
for computing a low rank approximation of a matrix, where $\ell_1$-norm relaxation is used to encourage sparsity.
In~\citep{d2007direct}, a convex relaxation for the $\ell_1$ constrained PCA is introduced and solved by semidefinite programming. In~\citep{journee2010generalized}, a generalized power (GPower) method is proposed for SPCA, where $\ell_0/\ell_1$ penalization is used.
In~\citep{luss2013conditional}, based on the well-known conditional gradient algorithm,
a framework called ConGradU is proposed, which unifies a variety of algorithms.
In~\citep{yuan2013truncated}, a truncated power method (TPower) is proposed,
which adopts the power method to update the approximate eigenvector,
followed by a truncation procedure that keeps a few largest magnitude entries of the approximate eigenvector
and truncates the remaining entries to zero. Also, see~\citep{d2008optimal, Proc:Moghaddam_ICML06} for other greedy algorithms proposed for  SPCA.

The SGEP, compared with SEP, is less investigated, especially for the large ill-conditioned problems.
SEP solvers such as GPower, ConGradU, and TPower only require matrix-vector product (MVP) operations and hence are efficient for large problems.
 For SGEP, however, those methods are no longer directly applicable.
In~\citep{sriperumbudur2011majorization}, SGEP is framed as a difference of convex functions program and solved via a sequence of convex programs
where the majorization-minimization method is used.
In~\citep{song2015sparse}, SGEP is transformed into a sequence of regular GEP via quadratic minorization functions,
and the preconditioned steepest ascent method is used to find the leading eigenpair.
In~\citep{safo2018sparse},
a general framework called sparse estimation with linear programming is proposed,
where the leading eigenpair of $(\wt{A},{\wt{B}})$ is used to simplify the constraint.
In~\citep{tan2018sparse}, the truncated Rayleigh flow method ({\sc rifle}) is proposed,
where the approximate eigenvector is updated via fixed step size steepest ascent method,
and followed by  simple truncation.

\textbf{Our proposal -- IFTRR.} We propose an inverse-free truncated Rayleigh-Ritz (IFTRR) method for SGEP. The classical Rayleigh-Ritz method is an approximate algorithm for computing eigenvalue equations~\citep{demmel1997applied,stewart2001matrix}. IFTRR  has three major steps:
first, the approximated eigenvector is updated via an inverse-free generalized eigensolver;
second, with the help of the updated eigenvector,
a truncation procedure is used to find the support set for the approximate eigenvector in the next iteration;
third, a small GEP is solved, and the approximate eigenvector is updated.

In the implementation of IFTRR, only matrix-vector product (MVP) is required, and hence the method is inherently suited for large scale problems. Furthermore, IFTRR is applicable for ill-conditioned or singular $\wt{A}$, $\wt{B}$. Additionally, the proposed truncation procedure, which is based on ``eigenvalue increment'' (see Section~\ref{sec:r3}),
is able to find the support set of the leading eigenvector effectively.
Our numerical experiments (Section~\ref{sec:numer}) show that IFTRR usually converges in a few iterations.

\vspace{0.1in}

\textbf{Notation.}
The symbol $\otimes$ denotes the Kronecker product.
The calligraphic letter $\I$ is usually used to denote an index set,
$|\I|$ denotes the cardinality of $\I$, e.g., $\I=\{i_1,i_2,\dots,i_s\}$,
where $i_1,i_2,\dots,i_s$ are distinct integers, then $|\I|=s$.
Let $a=[\alpha_1,\alpha_2,\dots,\alpha_p]^{\T}\in\R^p$, $A=[a_{jk}]\in\R^{p\times p}$,
 $a_{\I}$, $A_{\I}$ stand for $[\alpha_{i_1},\alpha_{i_2},\dots,\alpha_{i_s}]\in\R^{s}$
 and $[a_{i_ji_k}]\in\R^{s\times s}$, respectively.
For $w=[w_1,w_2,\dots,w_p]^{\T}\in\R^p$,
$\supp(w)$ denotes the index set of all nonzero entries of $w$,
$\supp(w,k)$ denotes the index set of the $k$ largest magnitude entries of $w$,
i.e., $\supp(w)\triangleq\{i\; | \; w_i\ne 0\}$, $\supp(w,k)\triangleq \{i_1,\dots,i_k \;|\; |w_{i_1}|\ge|w_{i_2}|\ge\dots\ge|w_{i_p}|\}$.
 $A_{(j,:)}$ and $A_{(:,k)}$ denote the $j$-th row and $k$-th column of $A$, respectively.
$I_p$ is the $p\times p$ identity matrix, and $e_j$ is its $j$-th column.
For symmetric definite matrix pairs $(A,B)$ and $(\wt{A},\wt{B})$,  we
denote $\rho(v)=\frac{v^{\T}Av}{v^{\T}Bv}$, $\tilde{\rho}(v)=\frac{v^{\T}\wt{A}v}{v^{\T}\wt{B}v}$.
The $i$th largest eigenvalue of $(A,B)$ is denoted by $\lambda_i(A,B)$.

\vspace{-0.05in}
\section{The Inverse Free Truncated Rayleigh-Ritz Method (IFTRR)}\label{sec:r3}
\vspace{-0.05in}

In this section, we present the inverse-free truncated Rayleigh-Ritz (IFTRR) method for solving the sparse generalized eigenvalue problem (SGEP). We first explain the intuition behind the development of the algorithm, before we present the details of the algorithm.

The mechanism of IFTRR is deceivingly simple: given an approximate eigenvector,
update the vector via certain eigensolvers,
then sparsify the resulting vector via a truncation procedure.

\vspace{-0.05in}
\subsection{Eigensolvers for Generalized  Eigenvalue Problem (GEP)}\label{eigensolver}
\vspace{-0.05in}

 For large scale generalized eigenvalue problem (GEP) $\wt{A}v=\lambda  \wt{B}v$,
where $\wt{A}$, $\wt{B}\in\mathbb{R}^{p\times p}$ are symmetric
and $\wt{B}$ is (semi) positive definite,
 iterative methods are usually used to compute its a few largest (or smallest) eigenvalues and the corresponding eigenvectors.
Simply speaking, these iterative methods  consist of two major steps:
 first, determine a subspace of $\mathbb{R}^p$;
 second, update the approximate eigenpairs via the Rayleigh-Ritz procedure.
Some detailed discussions follow.

\textbf{The subspace.}
The most popular subspace for the eigenvalue problem is the Krylov subspace.
Given a matrix $T\in\mathbb{R}^{p\times p}$ and a nonzero vector $v\in\mathbb{R}^{p}$, the order-$m$ Krylov subspace is defined as
\begin{align}\label{eqn:Kryylov}
\mathscr{K}_m(T, v)\triangleq\subspan\{v,Tv,T^2v,\dots,T^{m-1}v\}.
\end{align}
An orthonormal basis of $\mathscr{K}_m(T, v)$ can be obtained via the Arnoldi iteration,  e.g.,~\cite[Chapter 5]{stewart2001matrix},~\cite[Chapter 6]{demmel1997applied}.
Here we would like to emphasize that since the Arnoldi iteration only requires matrix vector product (MVP) $Tv$,
it is unnecessary to formulate $T$ explicitly and a subroutine that computes $Tv$ is sufficient.
For GEP, $T$ is usually set as $T=\wt{B}^{-1}\wt{A}$.
As $\wt{B}^{-1}$ is involved, the implementation of the MVP $Tv$ requires the MVP $u=\wt{A}v$ and also solving the linear system $\wt{B}z=u$ for $z$.
When the matrix size $p$ is large, it is expensive to solve the linear system $\wt{B}z=u$.
More importantly, when $\wt{B}$ is ill-conditioned (or even singular),
solving $\wt{B}z=u$ is not only expensive but also prone to large numerical errors.
In~\citep{golub2002inverse}, Golub and Ye propose to use $\mathscr{K}_m(\wt{A}-\rho \wt{B},v)$ to solve GEP,
where $\rho\in\R$ is a shift.
There are also other subspaces that can be used to solve GEP, e.g.,
the Davidson method~\citep{er1975iterative} and the Jacobi-Davidson method~\citep{sleijpen2000jacobi}.
In this paper, we use the Krylov subspace $\mathscr{K}_m(\wt{A}-\rho \wt{B},v)$, mainly due to its simplicity, scalability, and most importantly, it is inverse-free,
since only MVP $(\wt{A}-\rho \wt{B})v$ is required.

\textbf{The Rayleigh-Ritz procedure.}
Given an $m$-dimensional subspace $\mathcal{V}_m$ of $\mathbb{R}^p$ ($m\ll p$),
let $V_m$ be an orthonormal basis of $\mathcal{V}_m$.
The Rayleigh-Ritz procedure has three steps:
First, project the GEP $\wt{A}v=\lambda \wt{B}v$ onto $\mathcal{V}_m$, which yields a small GEP $(V^{\T}\wt{A}V)y=\mu (V^{\T}\wt{B}V) y$;
Second, solve the eigenpairs $(\mu_i,y_i)$ for $i=1,\dots,m$ of the small GEP;
Third, compute $(\tilde{\lambda}_i,\tilde{v}_i)=(\mu_i,V_m y_i)$ for $i=1,\dots,m$.
Then $(\tilde{\lambda}_i,\tilde{v}_i)$'s, often referred to as {\em Ritz pairs},
are used as approximate eigenpairs of the original  GEP $\wt{A}v=\lambda \wt{B}v$.

\vspace{-0.07in}
\subsection{The Truncation Procedure}\label{rt}
\vspace{-0.07in}

Let $(\rho,w)$ be a Ritz pair, which is used to approximate the leading eigenpair of $(\wt{A},\wt{B})$.
In general, $w$ is dense.
So, it is natural to sparsify $w$ since we are solving a sparse vector.
A popular way to accomplish this task is the so-called {\em truncation},
where the entries of $w$ are truncated to zeros except for the first $k$ largest magnitude entries.
However, this truncation procedure is found to be potentially misleading~\citep{cadima1995loading}.
In this paper, we propose to do  the ``truncation procedure'' as follows:
\begin{enumerate}\vspace{-0.1in}
\item 
Find a permutation $\{i_1,i_2,\dots,i_p\}$ of $\{1,2,\dots,p\}$ such that
$|w_{i_1}|\ge|w_{i_2}|\ge\dots\ge|w_{i_p}|$.
\item For $s=s_1,s_1+1,\dots, s_2$, set $\J=\{i_1,i_2,\dots,i_s\}$ and compute the leading eigenpair of the small GEP $(\wt{A}_{\J},\wt{B}_{\J})$, denoted by $(\rho_s,z_s)$,
$s_1<s_2$ are user-prescribed integers.
\item Determine the smallest $s$ such that $\rho_{s_2}-\rho_s\le \texttt{tol}$, where $\texttt{tol} > 0$ is a small real number.
\item Set $\I=\{i_1,i_2,\dots,i_s\}$, $\hat{v}_{\I}=z_s$ and $\hat{v}_{\I^c}=0$.
\end{enumerate}

The above truncation procedure is based on this  observation:
for any $\J\supseteq\I=\supp(v_1)$, we have
\begin{align*}
\lambda_1&=\max_{z\ne 0}\frac{z^{\T}A_{\I}z}{z^{\T}B_{\I}z}\le \max_{z\ne 0}\frac{z^{\T}A_{\J}z}{z^{\T}B_{\J}z}
\le \max_{z\ne 0}\frac{z^{\T}Az}{z^{\T}Bz}=\lambda_1.
\end{align*}
Thus, $\max_{z\ne 0}\frac{z^{\T}A_{\J}z}{z^{\T}B_{\J}z}\equiv \lambda_1$ as long as $\J\supseteq\I$.
In other words, when $\J$ is a superset of $\I$, the leading eigenvalue of $(A_{\J},B_{\J})$ remains a constant.
Therefore, for $(\wt{A},\wt{B})$, we also expect that when $\J$ contains $\I$ and $|\J|\ll p$,
the leading eigenvalue of $(\wt{A}_{\J}, \wt{B}_{\J})$ slightly changes.

Let $\hat{w}_{\I}=w$, $\hat{w}_{\I^c}=0$.
We prefer $\hat{v}$ rather than the simple truncated vector $\hat{w}$
simply because
\begin{align*}
\tilde{\rho}(\hat{v})=\max_{\supp(v)\subset\I} \tilde{\rho}(v) \ge \tilde{\rho}(\hat{w}),
\end{align*}
i.e., the target value of $\tilde{\rho}(v)$ at $v=\hat{v}$ is no less than that at $v=\hat{w}$.

In order to compare the simple truncation method with our ``eigenvalue increment'' method,
we take an approximate leading eigenvector $w=[w_1,\dots,w_p]^{\T}$ of $(\wt{A},\wt{B})$
from section~\ref{scca} ($p=1000$, $n=200$, $s=6$).
We sort the entries of $w$ such that $|w_{i_1}|\ge\dots\ge|w_{i_p}|$,
and compute $\rho_s=\lambda_1(\wt{A}_{\J_s},\wt{B}_{\J_s})$, where $\J_{s}=\{i_1,\dots,i_s\}$.

\begin{figure}[h!]
\begin{center}
\includegraphics[width=0.8\columnwidth]{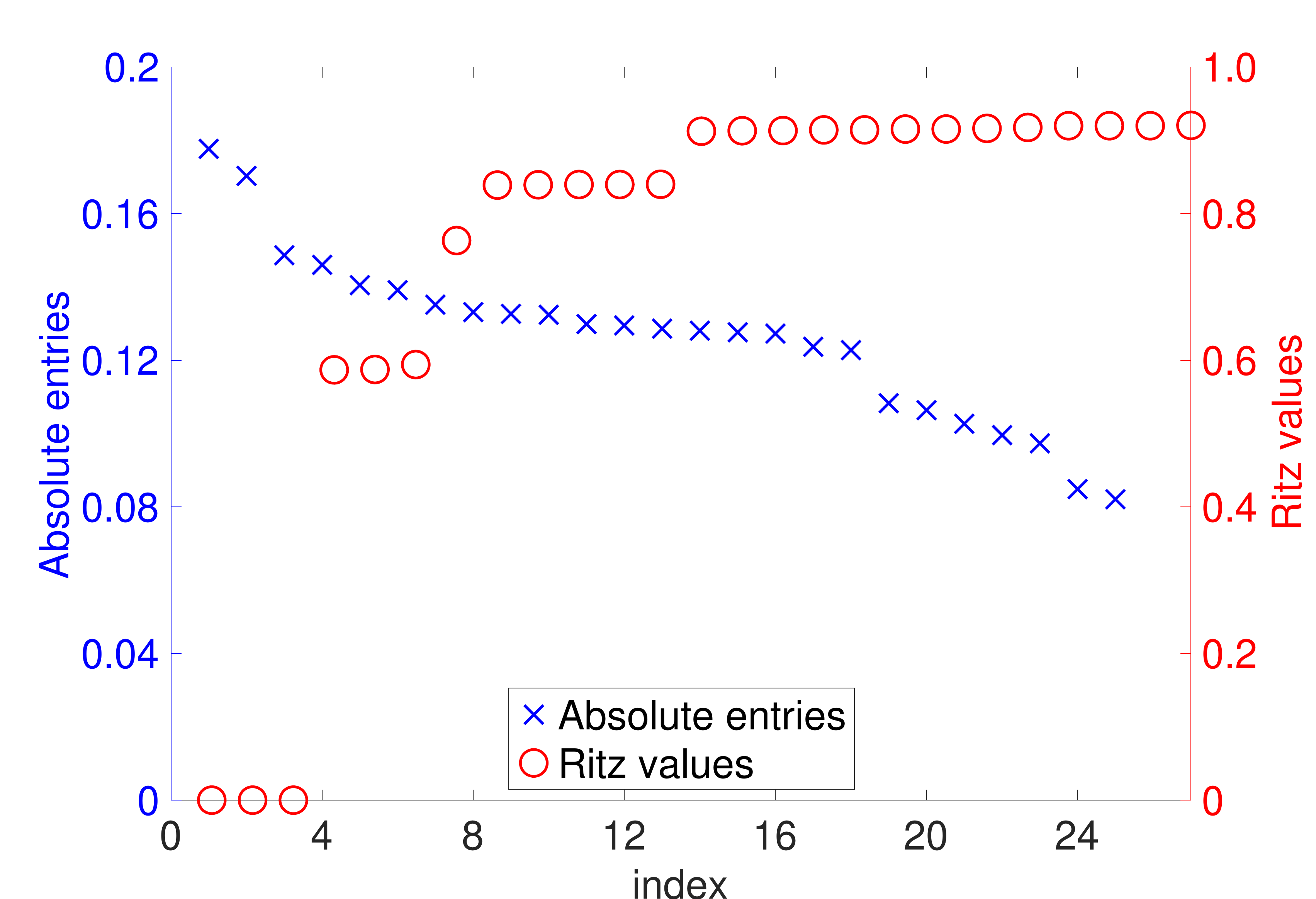}
\end{center}
\vspace{-0.2in}
\caption{Absolute entries vs.  eigenvalues.}
\label{fig:truncate}
\end{figure}

In Figure~\ref{fig:truncate}, we plot the top 25 entries $|w_{i_1}|,\dots,|w_{i_{25}}|$ and the 25 eigenvalues $\rho_1,\dots,\rho_{25}$.
We can see that there is no obvious gap among $|w_5|,\dots,|w_{18}|$, therefore,
it is difficult to determine a proper $k$ to do the simple truncation.
On the other hand, $\rho_{13},\dots,\rho_{25}$ almost remain unchanged,
then we may use $\I=\{i_1,\dots,i_{13}\}$ as the support set for a new approximate eigenvector.
In other words, $w_{i_{14}},\dots,w_{i_{1000}}$ are truncated to zeros.
As a matter of fact, for this example, the true support set of $v_1$ is contained in $\{i_1,\dots,i_{13}\}$.

The above example (and many others) 
show that our truncation procedure is effective to find the true support set of $v_1$.
As a result, the proposed algorithm usually converges in a few steps.

\vspace{-0.07in}
\subsection{Algorithm}\label{sec:alg}
\vspace{-0.07in}


We present the IFTRR method in Algorithm~\ref{alg:r3}.
Basically, IFTRR consists of three parts:
The first part (lines 5 to 7)
updates the leading eigenvector of $(\wt{A},\wt{B})$ via an eigensolver.
The second part determines an index set $\J$, which serves as the support of $v^{(t)}$ (lines 8 to 13),
then compute $v^{(t)}$ (line 17).
The third part (line 20) determines $\I$ and compute the final solution $\tilde{v}$ with $\supp(\tilde{v})=\I$.
The first two parts generate a sequence $\{(\rho^{(t)}, v^{(t)})\}_{t}$, and
the last part is the final update for the solution. Some implementation details and discussions of the algorithm follow.

\begin{algorithm}[htb]
   \caption{The proposed inverse-free truncated Rayleigh-Ritz (IFTRR) method. }
   \label{alg:r3}
\begin{algorithmic}[1]
   \State {\bfseries Input:} $\wt{A}$, $\wt{B}$, an integer $k$ for sparsity level, an integer $m$ for Krylov subspace dimension, and a randomly generated initial guess $v^{(0)}\in\R^{p}$. 
   \State {\bfseries Output:} An approximate solution $\tilde{v}$ to~\eqref{main}.
   \State Set $t=0$, $\rho^{(0)}=\tilde{\rho}(v^{(0)})$, $s_1=k$ and $s_2=k+\Delta k$; (e.g., $\Delta k$ = 20 or 30.)
   \While{unconverged and $t < \texttt{itermax}$}
   \State Compute an orthonormal $Q\in\R^{p\times m}$ such that $\subspan(Q)=\mathscr{K}_m(\wt{A}-\rho^{(t)}\wt{B},v^{(t)})$;
   \State Solve the leading eigenvector $\tilde{y}$ of $(Q^{\T}\wt{A}Q, Q^{\T}\wt{B}Q)$;
   \State Set $w=Q\tilde{y}$, $w=w/\|w\|_2$;
   \State Find a permutation $\{i_1,i_2,\dots,i_p\}$ of $\{1,2,\dots,p\}$ such that $|w_{i_1}|\ge |w_{i_2}|\ge \dots\ge|w_{i_p}|$;
   \For{$s=s_1,s_1+1,\dots,s_2$}
   \State Set $\J=\{i_1, i_2,\dots, i_s\}$;
   \State Solve the leading eigenpair of $(\wt{A}_{\J}, \wt{B}_{\J})$, denote it by $(\rho_s, z_s)$;
   \EndFor
   \State Find the smallest $s\in\{s_1,s_1+1,\dots,s_2\}$ such that $\rho_{s_2}-\rho_{s}\le (s_2-s)\times\texttt{tol}$;
   \State Set $\rho^{(t+1)}=\rho_s$, $v^{(t+1)}_{\J}=z_s/\|z_s\|_2$, $v^{(t+1)}_{\J^c}=0$, where $\J=\{i_1,i_2,\dots,i_s\}$;
   \State Set $t=t+1$;
   \EndWhile
   \State  Let $\I=\supp(v^{(t)},k)$,  solve the leading eigenvector $z$ of $(\wt{A}_{\I}, \wt{B}_{\I})$,
   set $\tilde{v}_{\I}=z/\|z\|_2$, $\tilde{v}_{\I^c}=0$.
\end{algorithmic}
\end{algorithm}

\textbf{Convergence test.}
The maximum number of iterations \texttt{itermax} is set as $100$.
The algorithm converges if one of the following conditions is satisfied:

(i) $\frac{\|(\wt{A}-\rho^{(t)} \wt{B})v^{(t)}\|_2}{\|\wt{A}\|_2 + |\rho^{(t)}| \|\wt{B}\|_2}<\texttt{tol}_1$,
where $\texttt{tol}_1$ is a tolerance and is set as $0.01$ in our  experiments.
This condition indicates that
$(\rho^{(t)},v^{(t)})$ is a good approximation of an eigenpair of $(\wt{A},\wt{B})$.

(ii) $|\rho^{(t)}-\rho^{(t-1)}|<\texttt{tol}_2$,
where $\texttt{tol}_2$ is a prescribed tolerance, say $10^{-3}$.
This condition indicates that the value of $\rho^{(t)}$ stagnates,
hence we may take  $\{\rho^{(t)}\}$ as a converged sequence.

\textbf{Solving a sequence of GEPs.}
At first glance, a sequence of generalized eigenvalue problems needs to be solved on lines 9 to 12.
But notice that
\[
\rho_{s+1}=\max_{z\ne 0}\frac{z^{\T}\wt{A}_{\J_2}z}{z^{\T}\wt{B}_{\J_2}z}
\ge \frac{\bsmat z_s\\ 0\esmat^{\T}\wt{A}_{\J_2}\bsmat z_s\\ 0\esmat}{\bsmat z_s\\ 0\esmat^{\T}\wt{B}_{\J_2}\bsmat z_s\\ 0\esmat}=\rho_s,
\]
where $\J_1=\{i_1,i_2,\dots, i_s\}$, $J_2=\{i_1,i_2,\dots, i_s, i_{s+1}\}$.
Thus $\{\rho_s\}_{s=s_1}^{s_2}$ is a non-decreasing sequence and we  use the idea of {\em bisection} to find the desired $s$ on line~13:
\\
\\ \mbox{}\hskip.2in
Set $a=s_1$, $b=s_2$;
\\ \mbox{}\hskip.2in
Compute $\rho_{a}$, $\rho_{b}$; 
\\ \mbox{}\hskip.2in
{\bf While} $b-a>1$ do
\\ \mbox{}\hskip.4in
 Set $s=\frac{a+b+\mod(a+b,2)}{2}$;
\\ \mbox{}\hskip.4in
 Compute $\rho_s$;
\\ \mbox{}\hskip.4in
If $\rho_{s_2}-\rho_s\le \texttt{tol}$, set $b=s$, $\rho_b=\rho_s$;
\\ \mbox{}\hskip.4in
Otherwise, set $a=s$, $\rho_a=\rho_s$.
\\ \mbox{}\hskip.2in
{\bf End while}

As a result, there are approximately $\log_2(s_2-s_1+1)$ small GEPs rather than $s_2-s_1+1$.

\textbf{Dealing with ill-conditioning.}
When $\wt{B}$ is singular, $\wt{B}_{\J}$ can also be singular.
As a result, infinity or indeterminate eigenvalues, which are sensitive to perturbations~\citep{bai2000templates}, occur,
then it will be difficult to determine which eigenpair is the leading one.
In our implementation, we use the following procedure as a cure for singular $\wt{B}_{\J}$:
First, compute the QR decomposition of $\wt{B}_{\J}$ with column pivoting~\cite[Chapter 5.4.2]{van2012matrix} $\wt{B}_{\J}\Pi=UR$,
where $\Pi$ is a permutation matrix, $U$ is an orthogonal matrix and $R$ is upper triangular with its diagonal entries non-negative and non-increasing;
Then let $\texttt{tol}_3>0$ be a user given threshold, say $\texttt{tol}_3=10^{-9}$.
Whenever $R_{(i,i)}<\texttt{tol}_3\times R_{(1,1)}$, we remove the corresponding index in $\J$.
For the resulting $\J$, $\wt{B}_{\J}$ is good conditioned.

\begin{table}[h!]
\vspace{-0.1in}
\caption{Computational complexity of Algorithm~\ref{alg:r3}. See the next paragraph ``\textbf{Computational complexity.} for the explanation of ``-''.}
\label{tab:r3}
\begin{center}
\begin{sc}
\begin{tabular}{ccc}
\hline
Line No. & Operation & Complexity \\
\hline
5 & mvp & -\\
5 & orthogonalization & $\mathcal{O}(m^2p)$\\
6 & eigenvalue prob. & $\mathcal{O}(m^3)$\\
8 & sorting & $\mathcal{O}(p\log p)$ \\
11, 17 & eigenvalue prob. & $\mathcal{O}(s^3)$\\
\hline
\end{tabular}
\end{sc}
\end{center}\vspace{-0.05in}
\end{table}

\textbf{Computational complexity.}
In Table~\ref{tab:r3}, we list the computational complexity of the major steps of the IFTRR method.
The symbol ``-'' means the computational complexity is different case by case:
(1) when $\wt{A}$ and $\wt{B}$ are available,
the operation $(\wt{A}-\rho^{(t)}\wt{B})v$ requires $\mathcal{O}(mp^2)$ FLOPS if $\wt{A}$ and $\wt{B}$ are dense,
$\mathcal{O}(m(\nnz(\wt{A})+\nnz(\wt{B})))$ FLOPS if $\wt{A}$ and $\wt{B}$ are sparse;
(2) when $\wt{A}$ and $\wt{B}$ are unavailable directly, the operations $\wt{A}v$ and $\wt{B}v$
are carried out via a data matrix $X\in\R^{n\times p}$, then the operation $(\wt{A}-\rho^{(t)}\wt{B})v$ in general
requires $\mathcal{O}(mnp)$ FLOPS if $\wt{A}$ and $\wt{B}$ are dense,
$\mathcal{O}(m \nnz({X}))$ FLOPS if $X$ are sparse.\footnote{For example, consider $X\in\R^{n\times p}$, where $n$ is the number of observations and $p$ is the number of features.
The sample correlation matrix is $C=\frac{1}{n} (X-{\bf 1}_n\bar{x}^{\T})^{\T}(X-{\bf 1}_n\bar{x}^{\T})$,
where ${\bf 1}_n=[1,\dots,1]^{\T}\in\R^n$, $\bar{x}=[\bar{x}_1,\dots,\bar{x}_p]^{\T}$,  $\bar{x}_i$ is the mean of the $i$-th column of $X$.
Then MVP $u=Cv$ is computed as $w=Xv- (\bar{x}^{\T} v){\bf 1}_n$, $u=\frac1n\left(X^{\T}w-({\bf 1}_n^{\T}w)\bar{x}\right)$,
which requires $\mathcal{O}(np)$ if $X$ is dense, and $\mathcal{O}(\nnz(X))$ if $X$ is sparse.
}
Since $s\ll p$ and $m\ll p$, the overall complexity of IFTRR is dominated by line 5.

\vspace{-0.07in}
\subsection{Convergence}
\vspace{-0.07in}

Before the study of the convergence, we give some definitions and preliminary lemmas.

The {\em angle} between two vectors $x$, $y\in\R^p$ is defined as
$\theta(x,y)\triangleq\arccos \frac{|x^{\T}y|}{\|x\|\|y\|}$.
Define the {\em Crawford number} for a definite-symmetric matrix pair $(A,B)$ as
\[
c(A,B)\triangleq \min_{\|x\|_2=1} \sqrt{(x^{\T} Ax)^2+(x^{\T} Bx)^2}.
\]
The following lemma tells that when $(A,B)$ is slightly perturbed, the changes of eigenvalues are small.
\begin{lemma}\cite[Theorem 8.7.3]{van2012matrix} \label{lem:ghe}
Suppose $(A,B)$ is a symmetric-definite pair with eigenvalues $\lambda_1\ge \lambda_2\ge \dots\ge \lambda_p$,  $E$ and $F$ are symmetric $p$-by-$p$ matrices that satisfy
\[
\epsilon=\sqrt{\|E\|_2^2+\|F\|_2^2} < c(A,B).
\]
Then $(A+E,B+F)$ is also a symmetric-definite pair with eigenvalues $\tilde{\lambda}_1\ge\tilde{\lambda}_2\ge \dots\ge \tilde{\lambda}_p$ that satisfy
\begin{align*}
|\arctan(\lambda_i)-\arctan(\tilde{\lambda}_i)|\le \arctan&(\epsilon/c(A,B)), \\
&\mbox{for}\quad i=1,2,\dots,p.
\end{align*}
\end{lemma}


Let $(\lambda^{(t)}, u^{(t)})$ be the current guess of the largest eigenpair of $(A,B)$ and  $Q_t\in\R^{p\times m}$ be an orthonormal basis for $\mathscr{K}_m(A-\lambda^{(t)}B, u^{(t)})$.
Then $(\lambda^{(t+1)},u^{(t+1)})$ can be obtained via the Rayleigh-Ritz procedure.
Specifically, let $(\omega,y)$ be the largest eigenpair of $(Q_t^{\T}AQ_t,Q_t^{\T}BQ_t)$,
then  $\lambda^{(t+1)}=\omega$, $u^{t+1}=Q_ty$.
By~\cite[Theorem 3.4]{golub2002inverse}, we have the following lemma.


\begin{lemma}\label{lem1}
Let the eigenvalues of $A-\lambda^{(t)}{B}$ be $\sigma_p\le \dots\le\sigma_2< \sigma_1$.
Assume $\lambda_2<\lambda^{(t)}<\lambda_1$.
Then
\begin{align*}
\lambda_1-\lambda^{(t+1)} \le ({\lambda}_1 - \lambda^{(t)})\epsilon_m^2
+\mathcal{O}(({\lambda}_1 - \rho^{(t)})^{\frac32}),
\end{align*}
where
$\epsilon_m=\min\limits_{p\in\mathcal{P}_m,\ p(\sigma_1)=1} \max\limits_{i\ne 1}|p(\sigma_i)|\le 2\Big(\frac{1-\sqrt{\psi}}{1+\sqrt{\psi}}\Big)^m$
with $\mathcal{P}_m$ denoting the set of all polynomials of degree not greater than $m$,
$\psi=\frac{\sigma_1-\sigma_2}{\sigma_1-\sigma_q}$.
\end{lemma}


Let $\rho^{(t)}$, $v^{(t)}$ be obtained via Algorithm~\ref{alg:r3}.
Denote  $\J_t=\supp(v^{(t)})$.
Here, we consider an alternative way to update $\rho^{(t)}$, $v^{(t)}$:

(i) Compute an orthonormal basis for $\mathscr{K}_m(\wt{A}_{\J_{t+1}}-\rho^{(t)}\wt{B}_{\J_{t+1}}, v^{(t)}_{\J_{t+1}})$, denote it as $Q_{\J_{t+1}}$;\\
(ii) Solve the largest eigenpair of $(Q_{\J_{t+1}}^{\T}\wt{A}_{\J_{t+1}}Q_{\J_{t+1}},Q_{\J_{t+1}}^{\T}\wt{B}_{\J_{t+1}}Q_{\J_{t+1}})$,
and denote it by $(\omega,y)$;\\
(iii) Set $\hat{\rho}^{(t+1)}=\omega$, $\hat{v}^{(t+1)}_{\J_{t+1}}=Q_{\J_{t+1}}y$, $\hat{v}^{(t+1)}_{\J_{t+1}^c}=0$.

Note that the above Rayleigh-Ritz procedure is only for the purpose of analyzing the convergence,
is not applicable in practice since $\J_{t+1}$ is unknown.
Also note that the existence of $Q_{\J_{t+1}}$ implicitly requires that
$m\le |\J_{t+1}|$,
meaning that {\em the dimension of the Krylov subspace should not exceed sparsity level of the approximate eigenvector}.

By Lemma~\ref{lem1}, we have the following result.

\vspace{-0.1in}

\begin{lemma}\label{lem2}
Let $\J_t=\supp(v^{(t)})$,
$\ell=|\J_{t+1}|$, the eigenvalues of $(\wt{A}_{\J_{t+1}},\wt{B}_{\J_{t+1}})$ be $\lambda_{1,{t+1}}\ge \dots\ge \lambda_{\ell,{t+1}}$,
the eigenvalues of $\wt{A}_{\J_{t+1}}-\rho^{(t)}\wt{B}_{\J_{t+1}}$ be $\sigma_{\ell}\le \dots\le\sigma_2< \sigma_1$.
Assume $\lambda_{2,{t+1}}<\rho^{(t)}<\lambda_{1,{t+1}}$.
Then
\begin{align*}
\lambda_{1,{t+1}}-\hat{\rho}^{(t+1)} \le ({\lambda}_{1,{t+1}} &- {\rho}^{(t)})\epsilon_m^2\\
&+\mathcal{O}(({\lambda}_{1,t+1} - {\rho}^{(t)})^{\frac32}),
\end{align*}
where $\epsilon_m$ is the same as in Lemma~\ref{lem1}.
\end{lemma}

Recall that $\rho^{(t+1)}$ is the largest eigenvalue of $(\wt{A}_{\J}, \wt{B}_{\J})$,
 $\hat{\rho}^{(t+1)}$ is the largest eigenvalue of $(Q_{\J_{t+1}}^{\T}\wt{A}_{\J_{t+1}}Q_{\J_{t+1}}, Q_{\J_{t+1}}^{\T}\wt{B}_{\J_{t+1}}Q_{\J_{t+1}})$, and $\J_{t+1}\subset\J$.
 Then we have

 \vspace{-0.1in}

\begin{lemma}\label{lem3}
It holds that
$\rho^{(t+1)}\ge \hat{\rho}^{(t+1)}$.
\end{lemma}


Define
\begin{subequations}
\begin{align}
&\eta_{s}^{(2)}\triangleq\max_{|\J|\le s}\lambda_2(\wt{A}_{\J},\wt{B}_{\J}),\\
&\eta_{s,\ell}^{(1)}\triangleq\max_{\stackrel{|\J|\le s}{|\J\cap\supp(v_1)|\le\ell}}\lambda_1(\wt{A}_{\J},\wt{B}_{\J}).
\end{align}
\end{subequations}
Combining Lemmas~\ref{lem2} and~\ref{lem3}, we have

\vspace{-0.1in}

\begin{theorem}\label{thm1}
Let $\J_t=\supp(v^{(t)})$, $s=\sup_t|\J_t|$ and $k=|\supp(v_1)|$.
For any $\J\subset[p]$ with $|\J|=s$,
denote the eigenvalues of $\wt{A}_{\J}-\rho^{(t)}\wt{B}_{\J}$ by $\sigma_{1,\J}>\sigma_{2,\J}\ge\dots\ge\lambda_{s,\J}$,
\[
\psi_*=\min_{|\J| = s} \frac{\sigma_{1,\J}-\sigma_{2,\J}}{\sigma_{1,\J}-\sigma_{s,\J}},\quad
\epsilon_*=2\Big(\frac{1-\sqrt{\psi_*}}{1+\sqrt{\psi_*}}\Big)^m.
\]

If $\eta_{s,k-1}^{(1)}\ge \rho^{(t)}>\eta_s^{(2)}$ and $|\J_t\cap\supp(v_1)|<k$,
then there exists a $\lambda_{1,t+1}\in(\rho^{(t)},+\infty)$ such that
\begin{align*}
\lambda_{1,{t+1}}-{\rho}^{(t+1)} \le ({\lambda}_{1,{t+1}} &- {\rho}^{(t)})\epsilon_*^2\\
&+\mathcal{O}(({\lambda}_{1,t+1} - {\rho}^{(t)})^{\frac32}).
\end{align*}
Asymptotically,
\[
\rho^{(t+1)}\gtrsim \rho^{(t)}+(\lambda_{1,t+1}-\rho^{(t)})(1-\epsilon_*^2).
\]
\end{theorem}

\vspace{-0.1in}

\begin{remark}
Let $s>k=|\supp(v_1)|$.
Assuming that for any $|\J|\le s$, $\|\wt{A}_{\J}-A_{\J}\|^2+\|\wt{B}_{\J}-B_{\J}\|^2$ is small,
by Lemma~\ref{lem:ghe}, we know that $\lambda_i(\wt{A}_{\J},\wt{B}_{\J})\approx \lambda_i({A}_{\J},{B}_{\J})$,
for all $i=1,2,\dots,s$.
By interlacing property (e.g.,~\cite[Theorem 8.1.7]{van2012matrix}), $\lambda_2({A}_{\J},{B}_{\J})\le \lambda_2$.
Therefore, we have
\begin{align*}
\eta_s^{(2)}\lesssim\lambda_2<\lambda_1\approx \eta_{s,k}^{(1)},
\end{align*}
i.e., the gap between $\eta^{(2)}_s$ and $\eta_{s,k}^{(1)}$ is larger than that between $\lambda_2$ and $\lambda_1$.
In fact,  in practice, the former is much larger than the latter.
\end{remark}

\vspace{-0.1in}

\begin{remark}
Let $k=|\supp(v_1)|$. Intuitively, we also expect a gap between $\eta_{s,k-1}^{(1)}$ and $\eta_{s,k}^{(1)}$,
the larger the gap is, the easier the problem is.
Otherwise, when the gap is sufficiently small, the problem has two ``solutions'':
one is approximately $v_1$, the other is $v'$ such that $\tilde{\rho}(v')=\eta^{(1)}_{s,k-1}$;
they are both sparse, and $\tilde{\rho}(v_1)\approx \tilde{\rho}(v')$.
In addition, notice that $\|v'\|_0=k-1< k=\|v_1\|_0$, consequently,
we can not expect to find good approximation of the true solution $v_1$, since $v'$ is a ``better'' solution,
in the sense that it is sparser and $\tilde{\rho}(v_1)\approx \tilde{\rho}(v')$.
\end{remark}

Theorem~\ref{thm1} tells that as long as $\supp(v_1)$ is not a subset of $\J_t$ and $\rho^{(t)}>\eta_s^{(2)}$,
$\{\rho^{(t)}\}_t$ is asymptotically nondecreasing.
Theorem~\ref{thm2} below tells that
once $\rho^{(t)}$ is larger than $\eta_{s,k-1}^{(1)}$, $\supp(v_1)$ is a subset of $\J_t$.


\begin{theorem}\label{thm2}
If $\rho^{(t)}>\eta^{(1)}_{s,k-1}$ with $k=|\supp(v_1)|$, then
\[
\supp(v_1)\subset\J_t.
\]
\end{theorem}


The following theorem tells that when $\supp(v_1)\subset\supp(v^{(t)})$,  then $\rho^{(t)}$ is close to $\lambda_1$ and $\theta(v^{(t)},v_1)$ is small, i.e., they are approximations of $\lambda_1$ and $v_1$.


\begin{theorem}\label{thm3}
Let $\J_t=\supp(v^{(t)})$.
Denote ${c}_{\J}=c({A}_{\J_t}, {B}_{\J_t})$,
$E_{\J}=\wt{A}_{\J_t}-A_{\J_t}$, $F_{\J}=\wt{B}_{\J_t}-B_{\J_t}$, and
${\epsilon}_{\J}=\sqrt{\|E_{\J}\|_2^2+ \|F_{\J}\|_2^2}$.
Assume $\supp(v_1)\subset\J_t$.

\noindent{\rm (a)} If ${\epsilon}_{\J}<{c}_{\J}$, then
\[
|\arctan(\rho^{(t)}) - \arctan(\lambda_1)|\le \arctan ({\epsilon}_{\J}/{c}_{\J}).
\]

\noindent{\rm (b)} Furthermore, if $|\rho^{(t)}| {\epsilon}_{\J} < {c}_{\J}$ and $\rho^{(t)}$ is simple,
then
\begin{align*}
\sin\theta(v^{(t)},v_1)\le \frac{\|\wt{B}_{\J_t}\|_2\delta+\sqrt{1+{\mu}^2}{\epsilon}_{\J}}{g}=\mathcal{O}({\epsilon}_{\J}),
\end{align*}
where 
$g$ is the smallest nonzero singular value of $\wt{A}_{\J_t}-\rho^{(t)}\wt{B}_{\J_t}$,
$\delta=\frac{(1+(\rho^{(t)})^2){\epsilon}_{\J}}{{c}_{\J}-|\rho^{(t)}|{\epsilon}_{\J}}$.
\end{theorem}

\vspace{-0.07in}
\section{Numerical Experiments}\label{sec:numer}
\vspace{-0.07in}

To illustrate the  behavior of the IFTRR method and compare it with existing methods for the SGEP, this section presents some numerical experiments.

\vspace{-0.07in}
\subsection{Sparse Canonical Correlation Analysis}\label{scca}
\vspace{-0.07in}

Recall Example~\ref{eg:cca}.
In our simulations, we set
$\Sigma_{xx}=\Sigma_{yy}=I_5\otimes D$, $D=[d_{jl}]\in\R^{p/10\times p/10}$ is a Toeplitz matrix with $d_{jl}=0.8^{|j-l|}$.
Let $v_x^*$ be collinear with $\sum_{j=1}^{s/2} e_{5j-4}$ and $(v_x^*)^{\T}\Sigma_{xx}v_x^*=1$,
where $s$ is a small even integer.
Set $v_y^*=v_x^*$,
$\Sigma_{xy}=0.9 \times \Sigma_{xx} v_x^* (v_y^*)^{\T}\Sigma_{yy}$ (low rank case), or
$\Sigma_{xy}=0.9 \times \Sigma_{xx} v_x^* (v_y^*)^{\T}\Sigma_{yy}+ 0.1 \times \Sigma_{xx} V_x^* (V_y^*)^{\T}\Sigma_{yy}$ (approximate low rank case),
where $V_x^*$, $V_y^*$ are random matrices such that $(V_x^*)^{\T}\Sigma_{xx}V_x^*=I_{p/2}$, $(V_y^*)^{\T}\Sigma_{yy}V_y^*=I_{p/2}$.

We perform the IFTRR method for 200 times under the following settings:\vspace{-0.1in}
\begin{enumerate}
\item Low rank $\Sigma_{xy}$, $p=1000$, $s=6$, different numbers of samples $n=100,200,300,400$;\vspace{-0.06in}
\item Low rank $\Sigma_{xy}$, $p=1000$, $n=400$, different numbers of sparsity levels $s=6,10,14,18$;\vspace{-0.06in}
\item Same as setting 1 except that $\Sigma_{xy}$ is approximate low rank;\vspace{-0.06in}  
\item Same as setting 2 except that $\Sigma_{xy}$ is approximate low rank. \vspace{-0.06in}
\item Low rank $\Sigma_{xy}$, $s=6$, $p=5000$, different numbers of samples $n=2000,4000,\dots,10000$;\vspace{-0.06in}
\item Low rank $\Sigma_{xy}$, $s=6$, $n=4000$, different numbers of features $p=2000,4000,\dots,10000$.\vspace{-0.06in}
\end{enumerate}

\begin{figure}[h!]
{\hspace{-0.3in}
\includegraphics[width=3.8in]{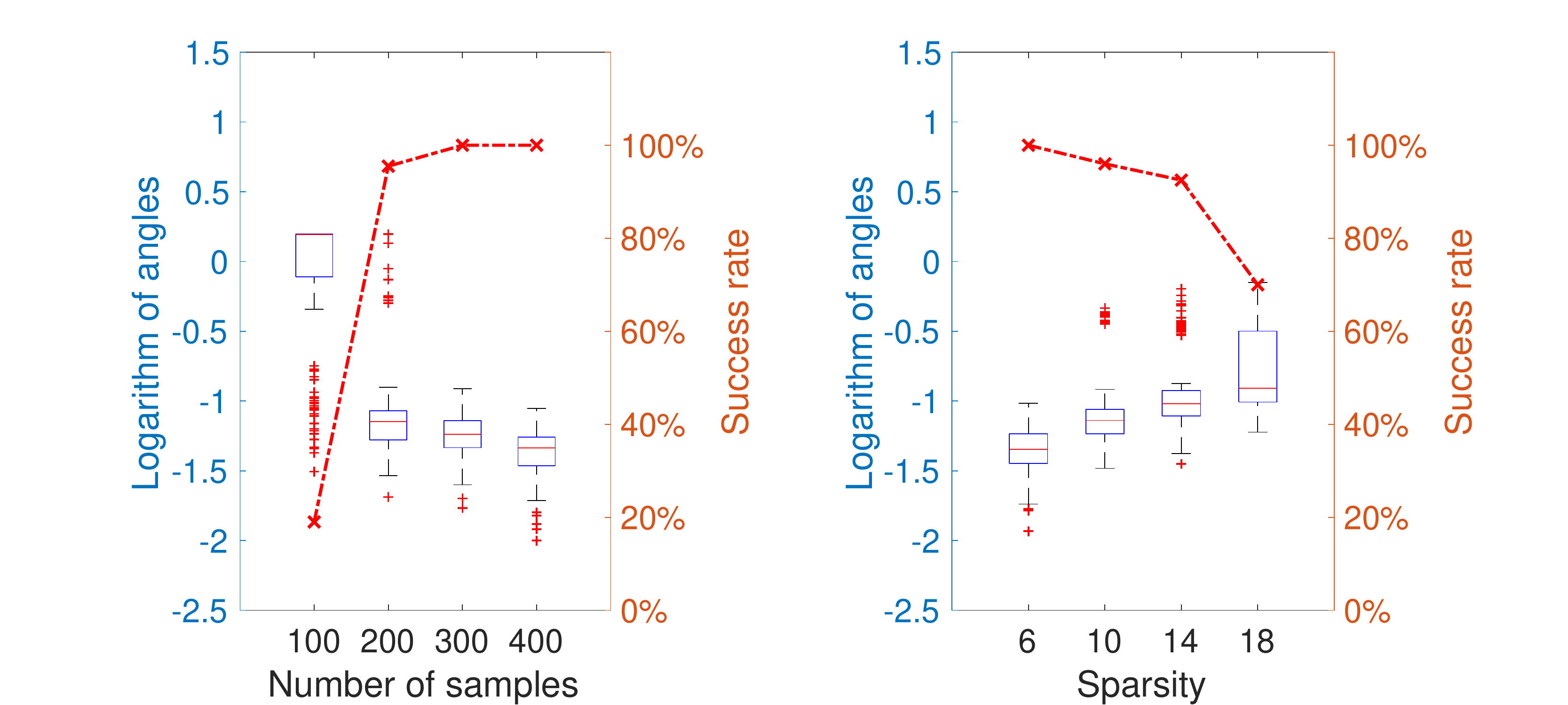}
}
\vspace{-0.25in}
\caption{Accuracy and success rate, low rank case, from left to right, settings 1 to 2.}
\label{fig:lowrank}
\vspace{-0.15in}
\end{figure}

\begin{figure}[h!]
{\hspace{-0.3in}
\includegraphics[width=3.8in]{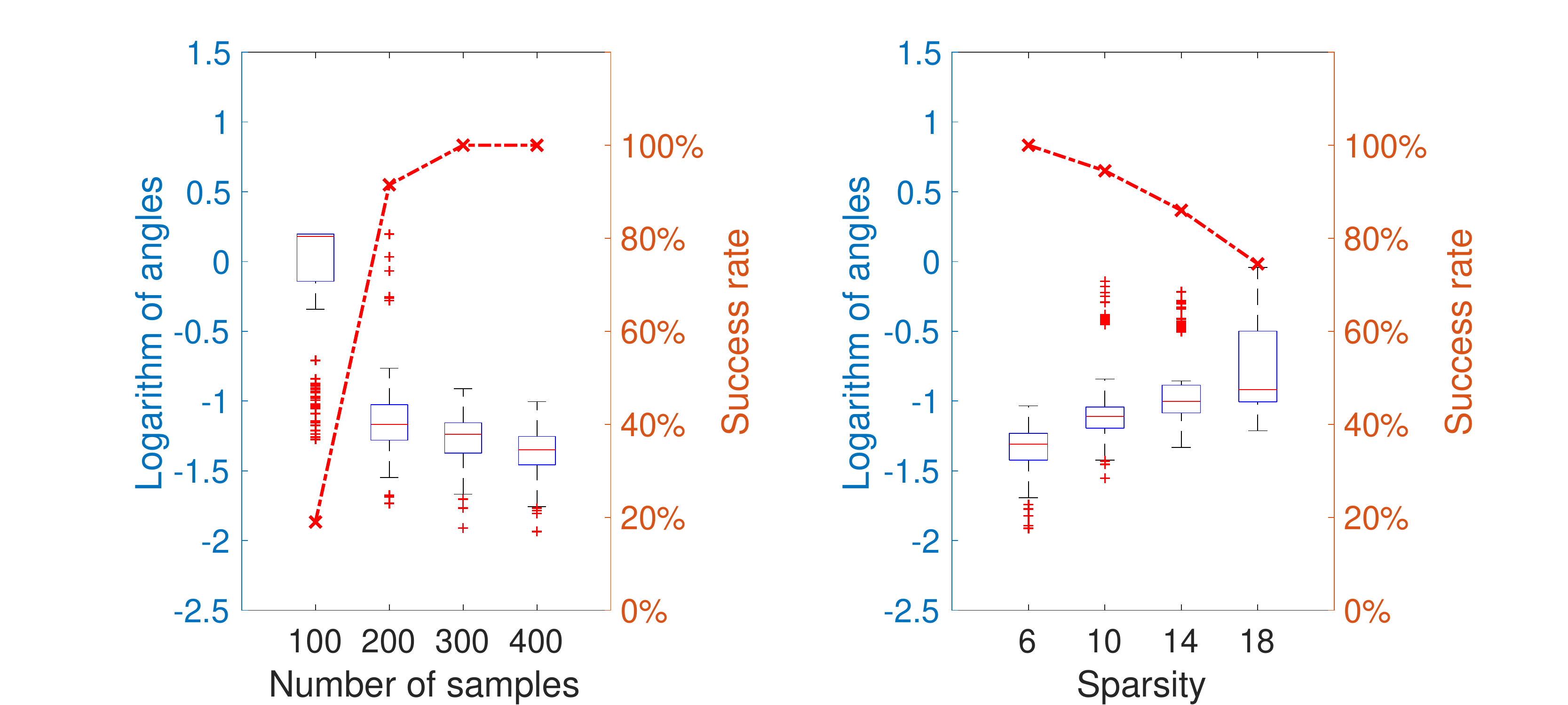}
}
\vspace{-0.25in}
\caption{Accuracy and success rate, approximate low rank case, from left to right, settings 3 to 4.}
\label{fig:applowrank}
\vspace{-0.15in}
\end{figure}

The performance of the method is evaluated in terms of the angle between $v_1=\bsmat v_x^*\\ v_y^*\esmat$ and the computed $\tilde{v}$,
and also success rate -- we say the returned $\tilde{v}$ is a success if $\supp(v_1)=\supp(\tilde{v})$.
The results are reported in Figures~\ref{fig:lowrank} and~\ref{fig:applowrank}.
We can see from these figures that
(i) for reasonable large $n$, $\tilde{v}$ (returned by the IFTRR method) is a good approximation of $v_1$,
the larger $n$ is, the smaller the angle is, and the larger the success rate is;
(ii) for different sparsity levels $s$, $\tilde{v}$ is also a good approximation of $v_1$,
the smaller $s$ is, the smaller the angle is, and the larger the success rate is;
(iii) the results for the low rank case are better than that for the approximate low rank case.
The above numerical results indicate that the IFTRR method gives a better result when the number of samples is sufficiently large and the leading eigenvector $v_1$ is sufficiently sparse.

\begin{figure}[h!]
\mbox{\hspace{-0.25in}
\includegraphics[width=1.95in]{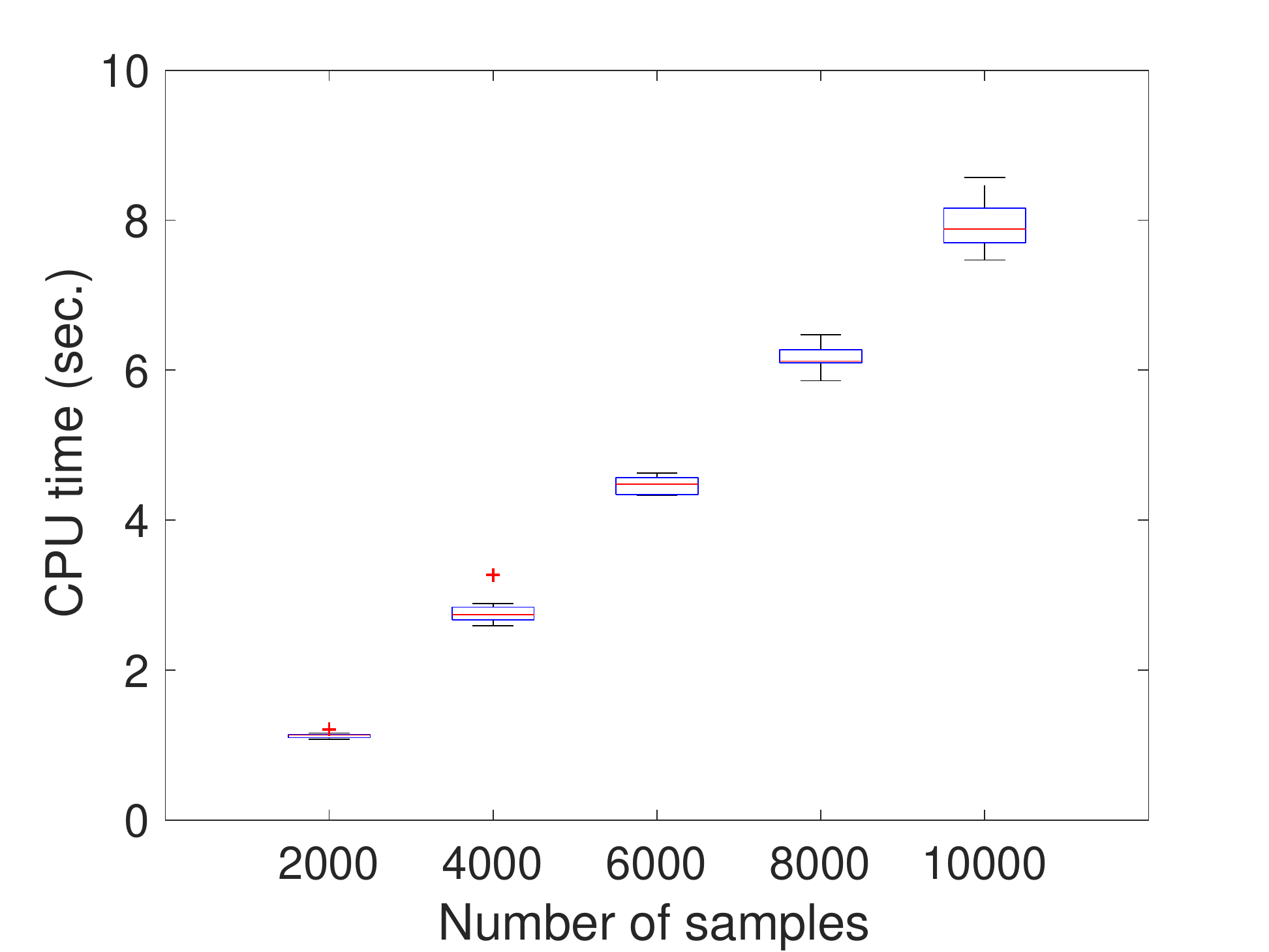}\hspace{-0.25in}
\includegraphics[width=1.95in]{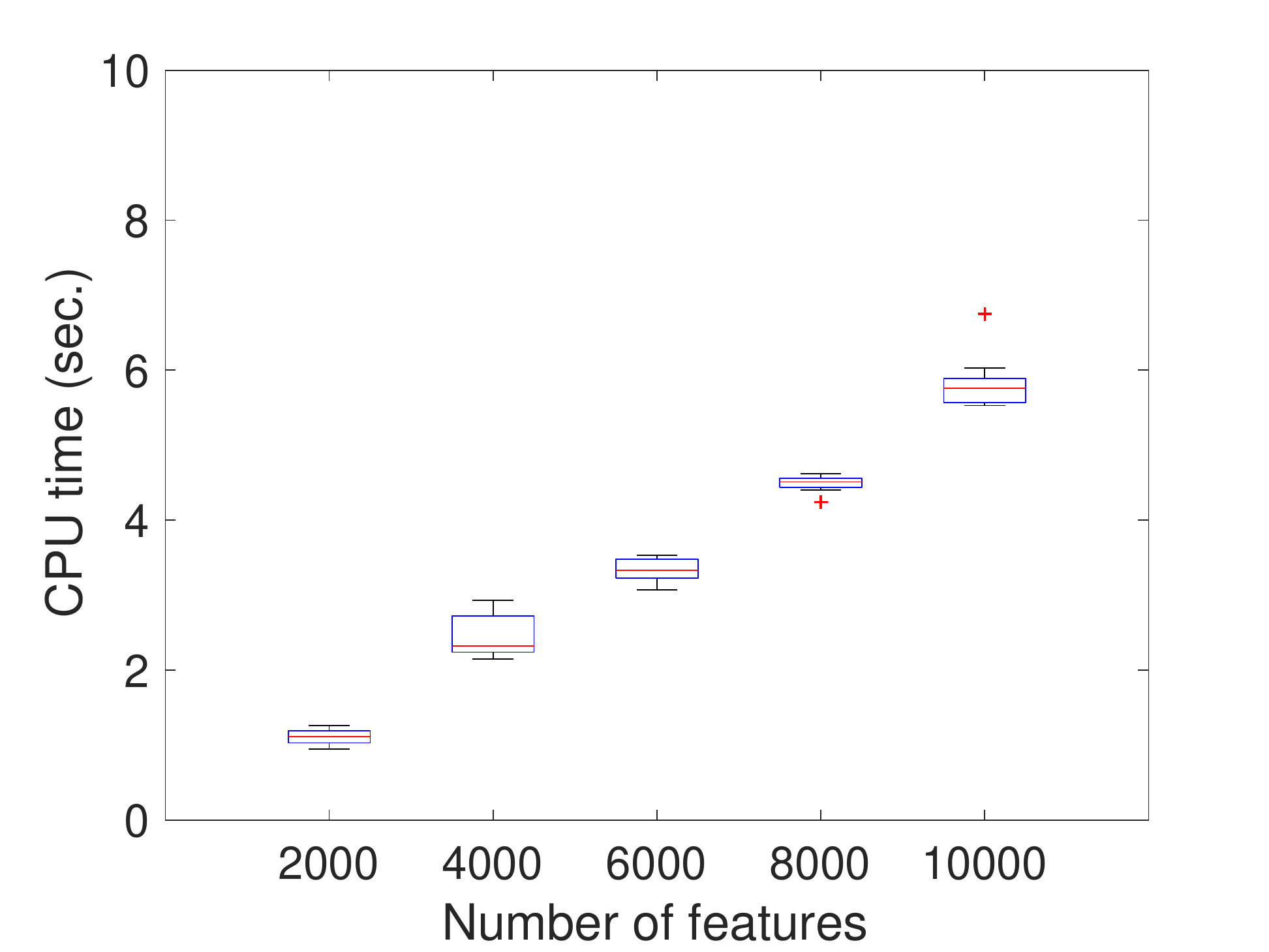}
}
\vspace{-0.25in}
\caption{CPU time, from left to right, settings 5 to 6.}
\label{fig:cputime}
\vspace{-0.15in}
\end{figure}

In Figure~\ref{fig:cputime}, we give the boxplots of the CPU time for settings 5 and 6.
We can see that with a fixed number of features,
the CPU time increases almost linearly with respect to the number of samples;
with a fixed number of samples,
the CPU time increases almost linearly with respect to the number of features.
This confirms  that the computational cost of the IFTRR method is dominated by MVP,
which is $\mathcal{O}(np)$.

\vspace{-0.07in}
\subsection{Sparse Fisher's Discriminant Analysis}\label{sfda}
\vspace{-0.07in}

Recall Example~\ref{eg:fda}.
In our simulations, for $k=1,2,\dots,K$, we set $\bar{x}_{kj}=\frac{2k-2}{K+2}$  for $j=2,4,\dots,40$, $\bar{x}_{kj}=0$ otherwise.
The data matrix $X$ is generated via $X_{(:,i)}\sim N(\bar{x}_k,\Sigma)$ for $i\in\mathcal{C}_k$,
where $\Sigma=I_5\otimes D$, $D=[d_{jl}]\in\R^{p/5\times p/5}$ is a Toeplitz matrix with $d_{jl}=0.8^{|j-l|}$.

\begin{table}[h!]
\caption{Results of Misclassification error and number of selected features.}
\label{tab:fda}
\vskip -0.4in
\begin{center}
\begin{small}
\begin{sc}
\begin{tabular}{cc|cccc}
\hline
&   $K$ & glmnet  & d(m)sda & rifle & IFTRR  \\
\hline
Err. & 2 & 32 & 29 & 15 & {\bf 14}(4)\\
Feat. & & 88& 105 & 42 & 42(1)\\
\hline
Err. & 4 & 495 & 247 & 192 & {\bf 103}(11)\\
Feat. & & 54 & 102 & 42 & 42(1)\\
\hline
\end{tabular}
\end{sc}
\end{small}
\end{center}
\end{table}

Fix $p=500$, for $K=2,4$, using 400 training samples,
we perform {\sc glmnet}~\citep{friedman2010regularization}, {\sc dsda/msda}~\citep{mai2015multiclass, mai2012direct}, {\sc rifle}~\citep{tan2018sparse} and the IFTRR method.
The results are then used to classify 1000 test samples, the misclassification error (denoted by {\sc Err.}) and number of selected features (denoted by {\sc Feat.}) are recorded.
{\sc Err.} and {\sc Feat.} are averaged over 200 independently generated datasets and reported in Table~\ref{tab:fda}.
There the number in the brackets is the corresponding standard error, and all numbers are rounded to the nearest integers.
Table~\ref{tab:fda} shows that for $K=2$, the misclassification errors of  {\sc rifle} and the IFTRR method are comparable and lower than the other two methods;
and for $K=4$, the IFTRR method has the lowest misclassification error.

\vspace{-0.07in}
\subsection{Sparse Sliced Inverse Regression}\label{fs}
\vspace{-0.07in}

Consider the sparse sliced inverse regression for the model
$Y=f(v_1^{\T}X,\dots,v_k^{\T}X,\epsilon)$,
where $Y$ is a univariate response, $X$ is $d$-dimensional covariates,
 $\epsilon$ is the stochastic error independent of $X$,
 $f$ is the link function, which is unknown.
 Under regularity conditions, the subspace spanned by $v_1,\dots,v_k$ can be identified via solving an SGEP with $\wt{A}=\wh{\Sigma}_{\mathbb{E}(X|Y)}$, $\wt{B}=\wh{\Sigma}_x$, where $\wh{\Sigma}_x$ is the sample covariance matrix of $X$,
  $\wh{\Sigma}_{\mathbb{E}(X|Y)}$ is the sample covariance matrix of the conditional expectation $\mathbb{E}(X|Y)$, which is
$\wh{\Sigma}_{\mathbb{E}(X|Y)}=\wh{\Sigma}_x-\frac{1}{n_1+n_2} \sum_{k=1}^2 n_k \wh{\Sigma}_{x,k}$,
where $n_k$ is the number of samples for class $k$,
$\wh{\Sigma}_{x,k}$ is the sample covariance matrix for class $k$, for $k=1,2$.
See~\citep{chen2010coordinate,li1991sliced,li2007sparse,tan2018sparse} and reference therein for more details.

\begin{figure}[h!]
\mbox{\hspace{-0.25in}
\includegraphics[width=1.88in]{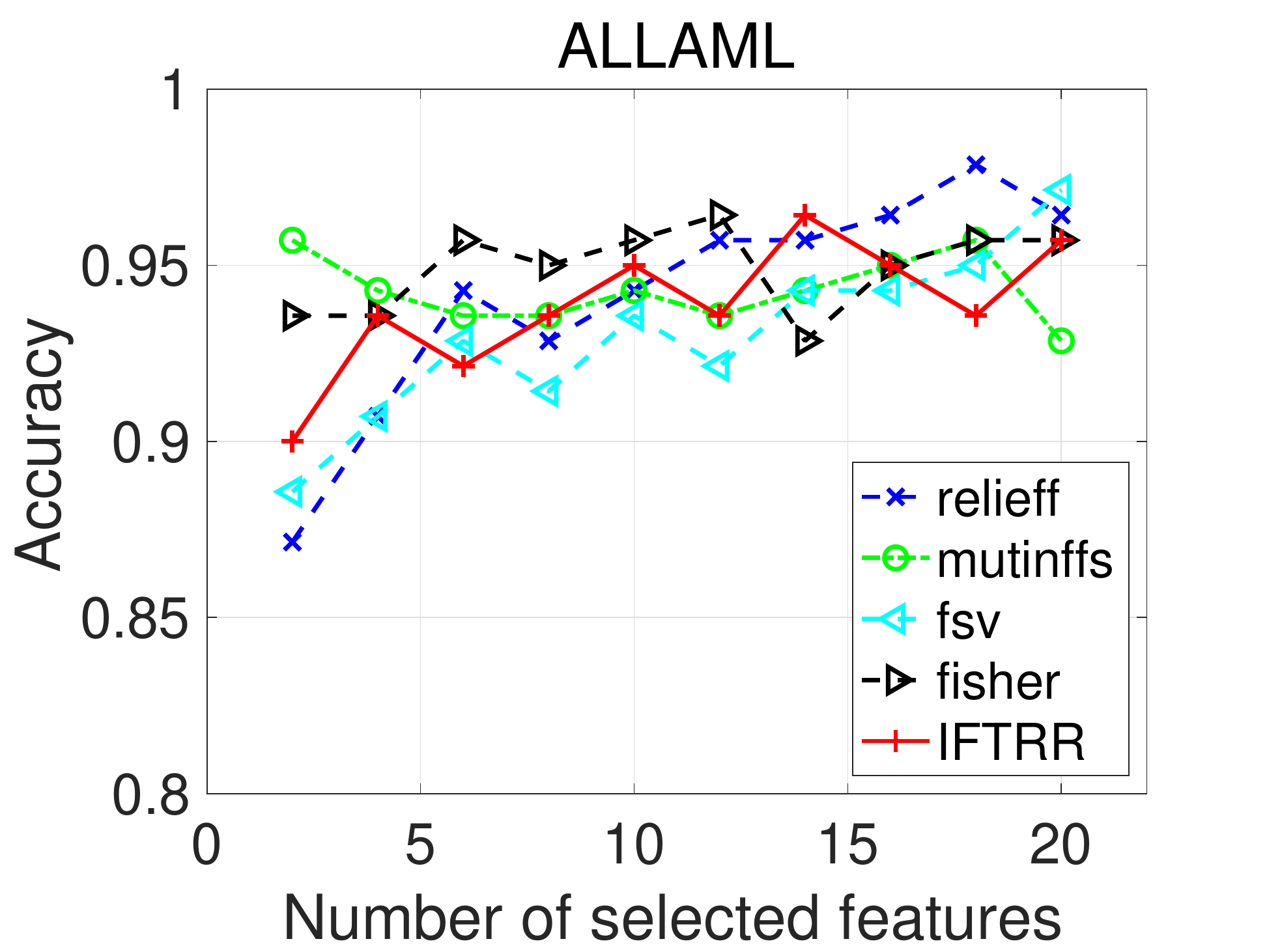}\hspace{-0.2in}
\includegraphics[width=1.88in]{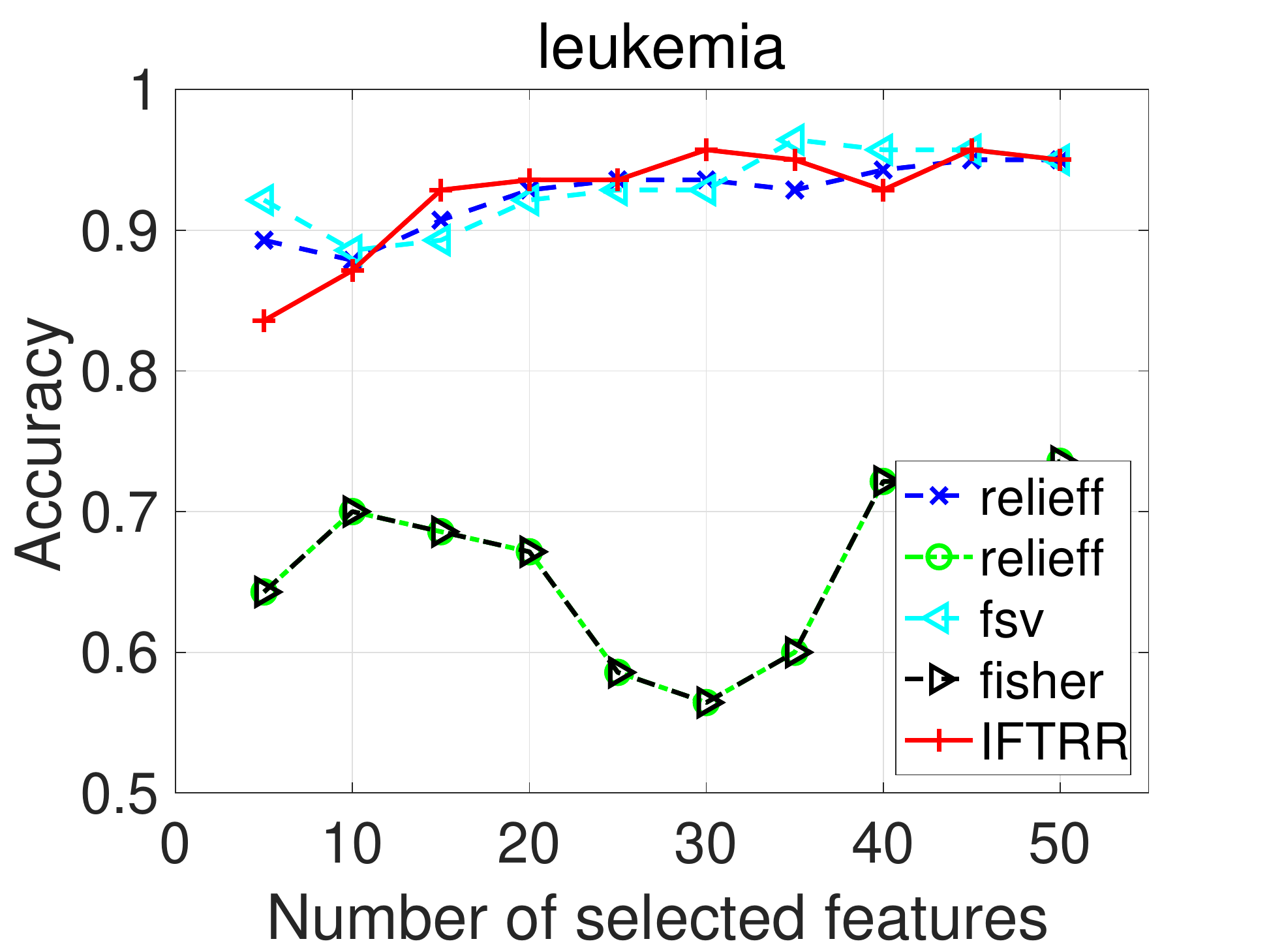}
}
\\[0.05in]
\mbox{\hspace{-0.25in}
\includegraphics[width=1.88in]{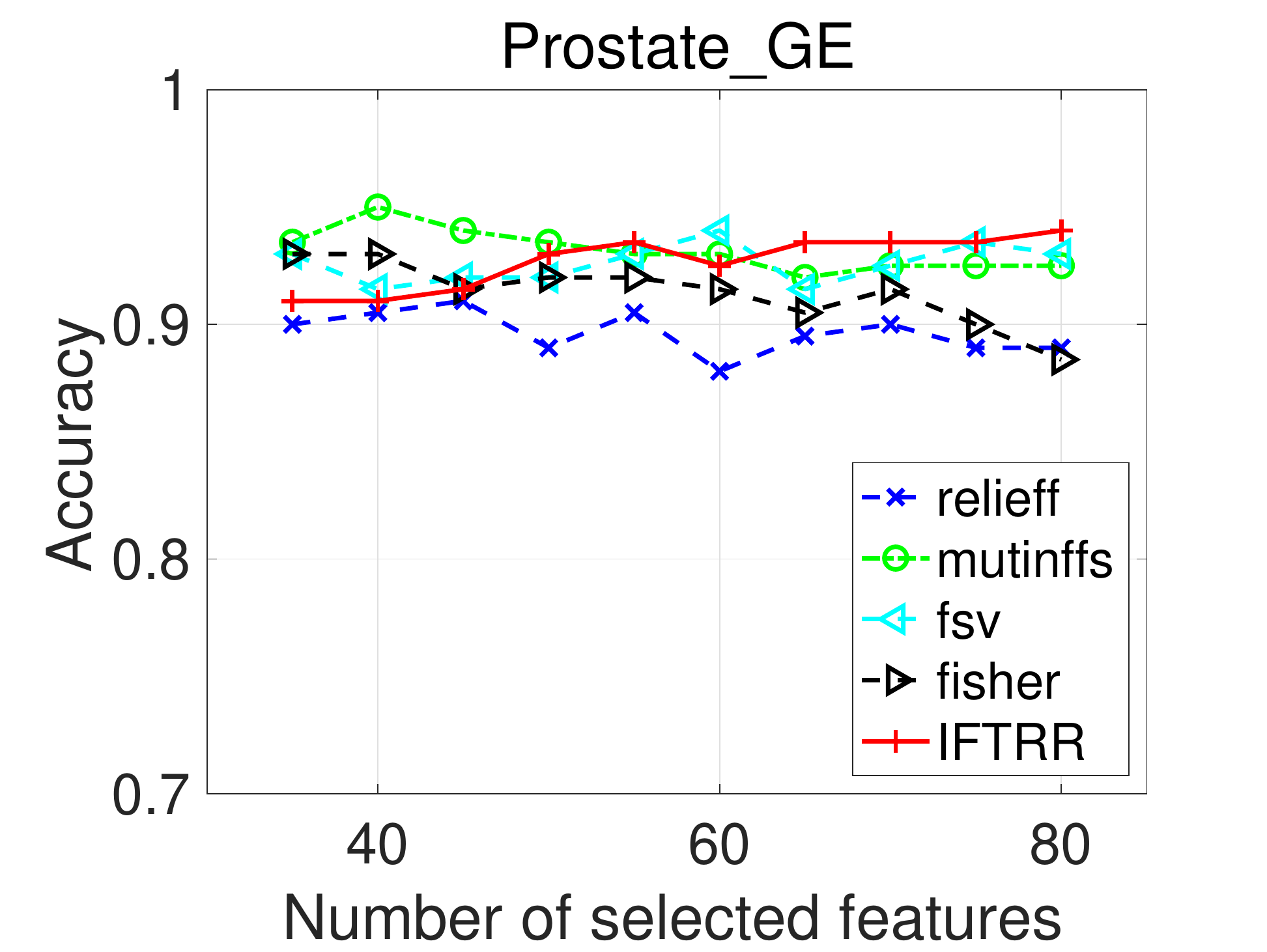}\hspace{-0.2in}
\includegraphics[width=1.88in]{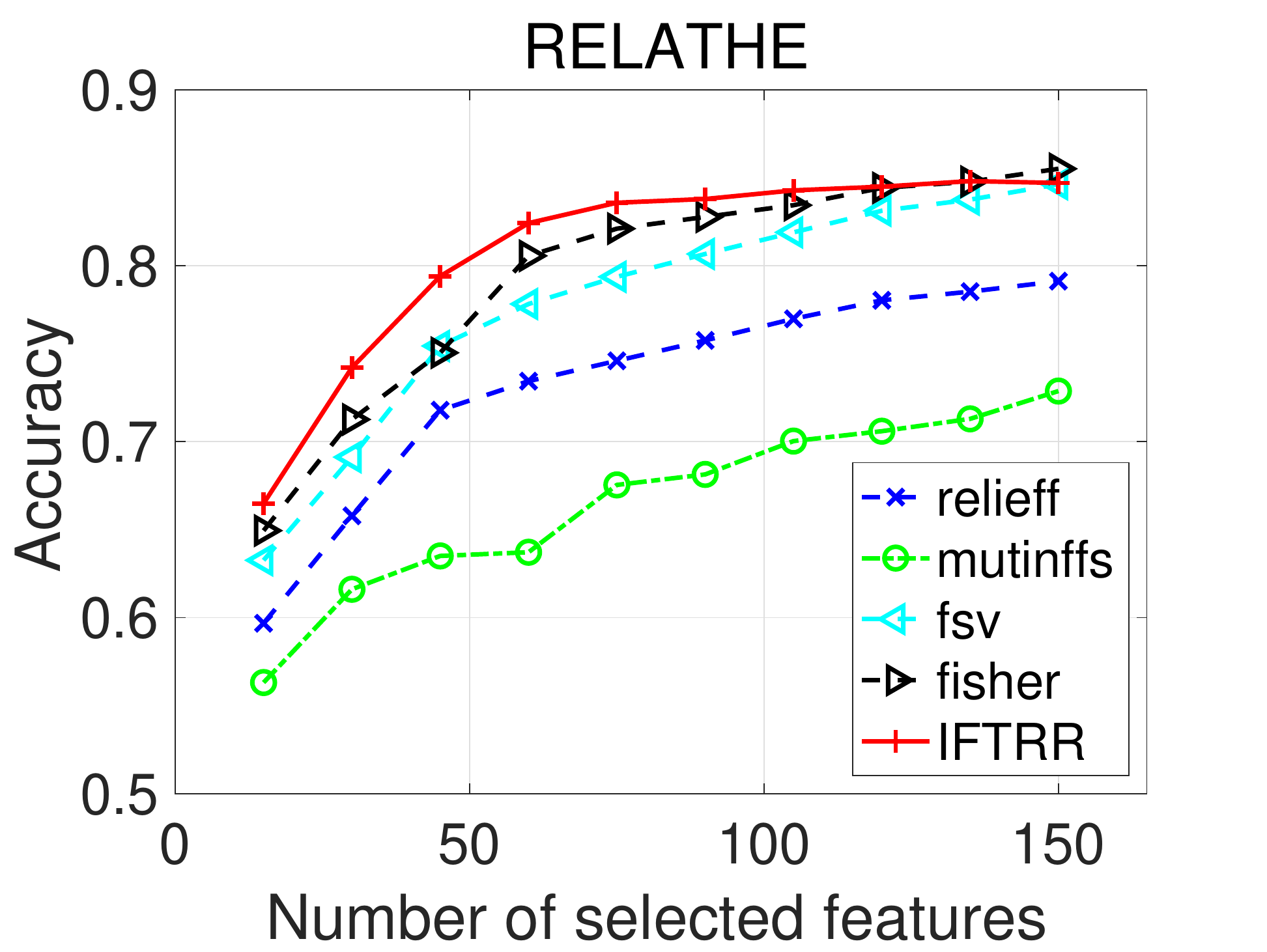}
}
\\[0.05in]
\mbox{\hspace{-0.25in}
\includegraphics[width=1.88in]{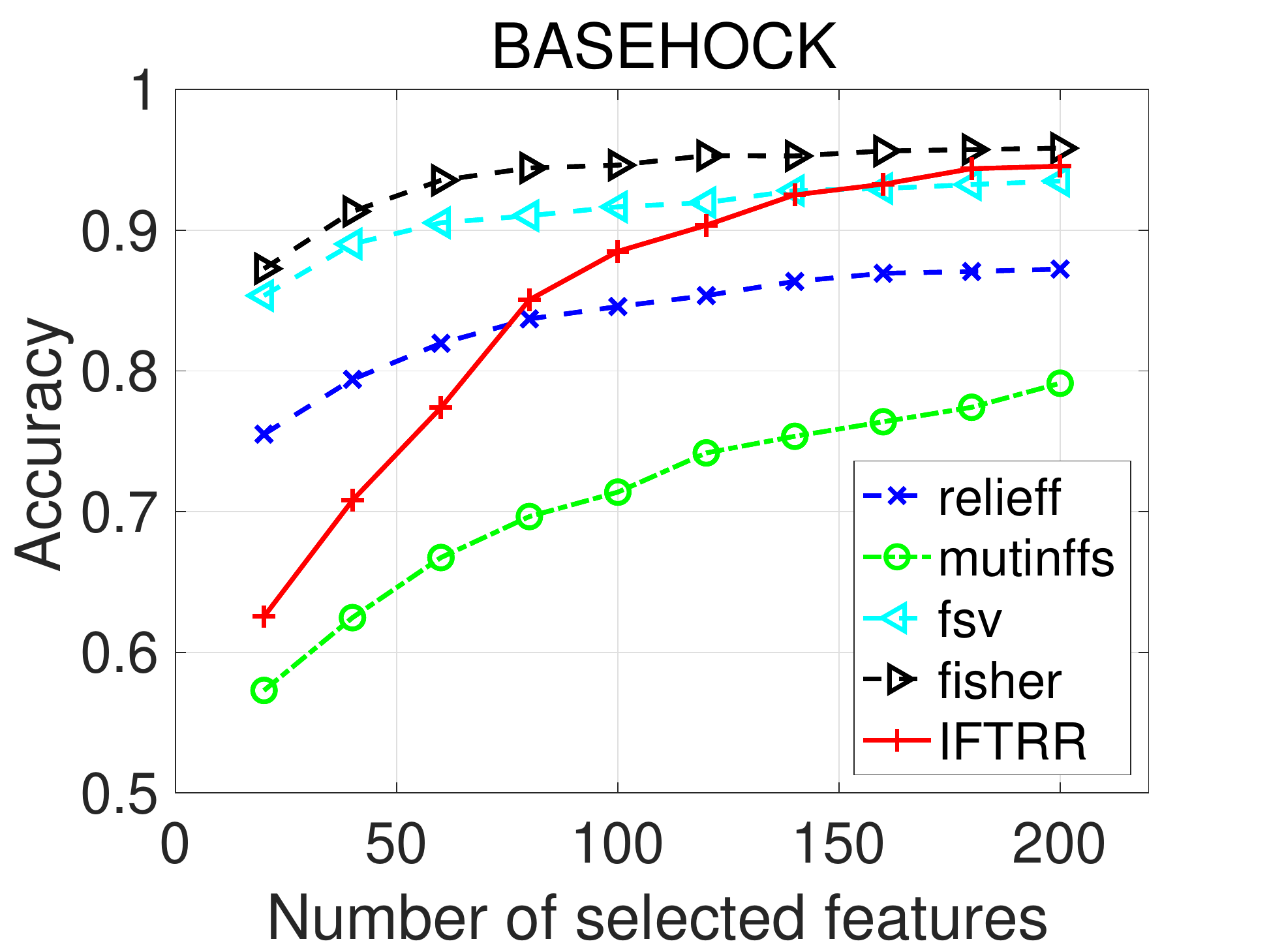}\hspace{-0.2in}
\includegraphics[width=1.88in]{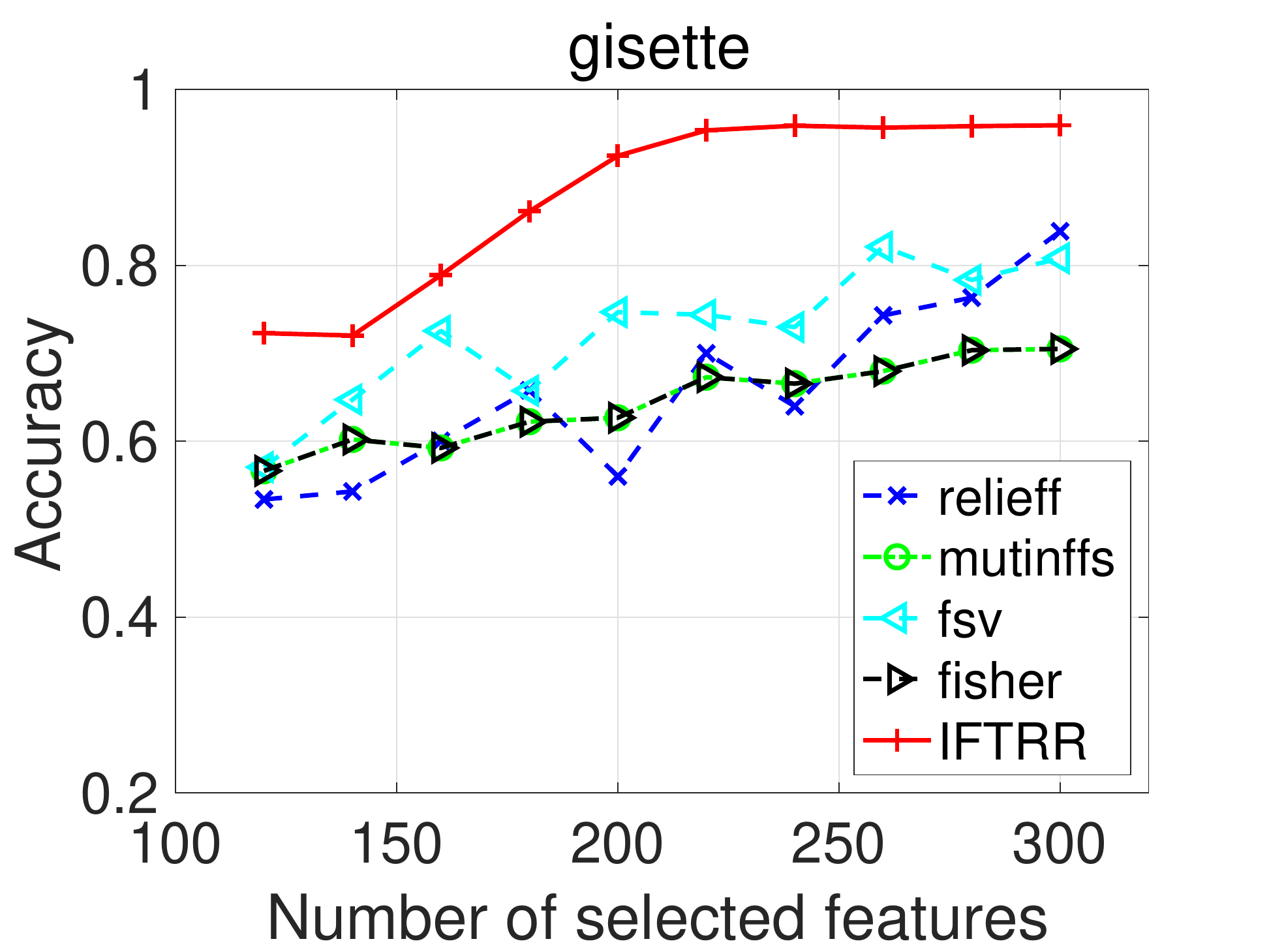}
}
\vspace{-0.2in}
\caption{Accuracy vs. number of features.}
\label{fig:acc}\vspace{-0.1in}
\end{figure}

Let $v^{(t)}$ be an approximate solution to SGEP.
$Xv^{(t)}$ is usually used as a predictor.
Here we may also use the indices for the nonzero entries of $v^{(t)}$ to select the features, and the features can be ranked by ordering the absolute values of the nonzero entries of $v^{(t)}$.
Now we compare our method with feature selection methods -- relieff, mutinffs, fsv and fisher, which are all available in the Feature Selection Library~\citep{roffo2017ranking,roffo2016ranking,Proc:Roffo_ICCV17,Proc:Roffo_ICCV15}. 
The datasets are all downloaded from scikit-feature feature selection repository.
Each dataset is randomly partitioned into a training set and a test set, and the test set size is approximately $0.2n$.
A support vector machine (SVM) classifier is trained using the training set with only the selected features and then used to predict on the test set.
The average accuracy (averaged over 10 independent runs) is plotted in Figure~\ref{fig:acc}.
We can see that our method is comparable with the other four methods for the first 5 datasets, outperforms the other four methods for the last dataset.

\section{Conclusion}\label{sec:conclusion}
\vspace{-0.08in}

We have proposed the IFTRR method to solve the SGEP.
The IFTRR method has the following advantages:
Since only the  MVP is required, the method is suited for large scale problem;
A cure is incorporated into the IFTRR method, which makes it applicable for ill-conditioned or singular coefficient matrices $\wt{A}$, $\wt{B}$;
Based on ``eigenvalue increment'', a new truncation procedure is proposed,
which is able to find the support set of the leading eigenvector effectively,
as a result, the IFTRR method usually converges in a few iterations.
Numerical simulations show that the IFTRR method is effective and efficient,
especially when the matrix size is large and the leading eigenvector is very sparse.

There are several future research topics for the IFTRR method.
First, can we extend the IFTRR method to compute multiple leading sparse eigenvectors?
Computing several leading eigenvectors one by one seems simple,
how to compute them simultaneously is uneasy
since the orthogonalization procedure, which is required for computing several eigenvectors simultaneously,
usually destroys the sparsity.
Second, how to truncate the eigenvector to ensure some structured sparsity, say group sparsity as in the group LASSO.
Third, based on the IFTRR method, a general framework for solving SGEP can be obtained --
{\tt eigensolver} + {\tt truncation}.
Various eigensolvers together with certain truncation procedure
can be tried to solve the SGEP.
Which is the best choice, how is the convergence?
More studies towards these directions are apparently required.

%
%
%

\clearpage

\balance
\bibliographystyle{plainnat}
\bibliography{myref,standard}

\begin{thebibliography}{33}
\providecommand{\natexlab}[1]{#1}
\providecommand{\url}[1]{\texttt{#1}}
\expandafter\ifx\csname urlstyle\endcsname\relax
  \providecommand{\doi}[1]{doi: #1}\else
  \providecommand{\doi}{doi: \begingroup \urlstyle{rm}\Url}\fi

\bibitem[Bai et~al.(2000)Bai, Demmel, Dongarra, Ruhe, and van~der
  Vorst]{bai2000templates}
Zhaojun Bai, James Demmel, Jack Dongarra, Axel Ruhe, and Henk van~der Vorst.
\newblock \emph{Templates for the solution of algebraic eigenvalue problems: a
  practical guide}.
\newblock SIAM, 2000.

\bibitem[Cadima and Jolliffe(1995)]{cadima1995loading}
Jorge Cadima and Ian~T Jolliffe.
\newblock Loading and correlations in the interpretation of principle
  compenents.
\newblock \emph{J. Appl. Stat.}, 22\penalty0 (2):\penalty0 203--214, 1995.

\bibitem[Chen et~al.(2010)Chen, Zou, Cook, et~al.]{chen2010coordinate}
Xin Chen, Changliang Zou, R~Dennis Cook, et~al.
\newblock Coordinate-independent sparse sufficient dimension reduction and
  variable selection.
\newblock \emph{The Annals of Statistics}, 38\penalty0 (6):\penalty0
  3696--3723, 2010.

\bibitem[Clemmensen et~al.(2011)Clemmensen, Hastie, Witten, and
  Ersb{\o}ll]{clemmensen2011sparse}
Line Clemmensen, Trevor Hastie, Daniela Witten, and Bjarne Ersb{\o}ll.
\newblock Sparse discriminant analysis.
\newblock \emph{Technometrics}, 53\penalty0 (4):\penalty0 406--413, 2011.

\bibitem[d'Aspremont et~al.(2007)d'Aspremont, El~Ghaoui, Jordan, and
  Lanckriet]{d2007direct}
A.~d'Aspremont, L.~El~Ghaoui, M.~Jordan, and G.~Lanckriet.
\newblock A direct formulation for sparse {PCA} using semidefinite programming.
\newblock \emph{SIAM Rev.}, 49\penalty0 (3):\penalty0 434--448, 2007.

\bibitem[Davidson(1975)]{er1975iterative}
Ernest~R. Davidson.
\newblock The iterative calculation of a few of the lowest eigenvalues and
  corresponding eigenvectors of large real-symmetric matrices.
\newblock \emph{J. Comput. Phys.}, 17:\penalty0 87--94, 1975.

\bibitem[Demmel(1997)]{demmel1997applied}
James~W Demmel.
\newblock \emph{Applied Numerical Linear Algebra}.
\newblock SIAM, Philadelphia, PA, 1997.

\bibitem[d’Aspremont et~al.(2008)d’Aspremont, Bach, and
  Ghaoui]{d2008optimal}
Alexandre d’Aspremont, Francis Bach, and Laurent~El Ghaoui.
\newblock Optimal solutions for sparse principal component analysis.
\newblock \emph{J. Machine Learning Res.}, 9\penalty0 (Jul):\penalty0
  1269--1294, 2008.

\bibitem[Friedman et~al.(2010)Friedman, Hastie, and
  Tibshirani]{friedman2010regularization}
Jerome Friedman, Trevor Hastie, and Rob Tibshirani.
\newblock Regularization paths for generalized linear models via coordinate
  descent.
\newblock \emph{Journal of statistical software}, 33\penalty0 (1):\penalty0 1,
  2010.

\bibitem[Golub and Ye(2002)]{golub2002inverse}
Gene~H Golub and Qiang Ye.
\newblock An inverse free preconditioned krylov subspace method for symmetric
  generalized eigenvalue problems.
\newblock \emph{SIAM J. Sci. Comput.}, 24\penalty0 (1):\penalty0 312--334,
  2002.

\bibitem[Jolliffe et~al.(2003)Jolliffe, Trendafilov, and
  Uddin]{jolliffe2003modified}
Ian~T Jolliffe, Nickolay~T Trendafilov, and Mudassir Uddin.
\newblock A modified principal component technique based on the {LASSO}.
\newblock \emph{J. Comput. Graphical Statist.}, 12\penalty0 (3):\penalty0
  531--547, 2003.

\bibitem[Journ{\'e}e et~al.(2010)Journ{\'e}e, Nesterov, Richt{\'a}rik, and
  Sepulchre]{journee2010generalized}
Michel Journ{\'e}e, Yurii Nesterov, Peter Richt{\'a}rik, and Rodolphe
  Sepulchre.
\newblock Generalized power method for sparse principal component analysis.
\newblock \emph{J. Machine Learning Res.}, 11\penalty0 (Feb):\penalty0
  517--553, 2010.

\bibitem[Li(1991)]{li1991sliced}
Ker-Chau Li.
\newblock Sliced inverse regression for dimension reduction.
\newblock \emph{J. Am. Stat. Assoc.}, 86\penalty0 (414):\penalty0 316--327,
  1991.

\bibitem[Li(2007)]{li2007sparse}
Lexin Li.
\newblock Sparse sufficient dimension reduction.
\newblock \emph{Biometrika}, 94\penalty0 (3):\penalty0 603--613, 2007.

\bibitem[Luss and Teboulle(2013)]{luss2013conditional}
Ronny Luss and Marc Teboulle.
\newblock Conditional gradient algorithmsfor rank-one matrix approximations
  with a sparsity constraint.
\newblock \emph{SIAM Rev.}, 55\penalty0 (1):\penalty0 65--98, 2013.

\bibitem[Mai et~al.(2012)Mai, Zou, and Yuan]{mai2012direct}
Qing Mai, Hui Zou, and Ming Yuan.
\newblock A direct approach to sparse discriminant analysis in ultra-high
  dimensions.
\newblock \emph{Biometrika}, 99\penalty0 (1):\penalty0 29--42, 2012.

\bibitem[Mai et~al.(2015)Mai, Yang, and Zou]{mai2015multiclass}
Qing Mai, Yi~Yang, and Hui Zou.
\newblock Multiclass sparse discriminant analysis.
\newblock \emph{arXiv:1504.05845}, 2015.

\bibitem[Moghaddam et~al.(2005)Moghaddam, Weiss, and
  Avidan]{Proc:Moghaddam_NIPS05}
Baback Moghaddam, Yair Weiss, and Shai Avidan.
\newblock Spectral bounds for sparse {PCA:} exact and greedy algorithms.
\newblock In \emph{Advances in Neural Information Processing Systems (NIPS)},
  pages 915--922, Vancouver, Canada, 2005.

\bibitem[Moghaddam et~al.(2006)Moghaddam, Weiss, and
  Avidan]{Proc:Moghaddam_ICML06}
Baback Moghaddam, Yair Weiss, and Shai Avidan.
\newblock Generalized spectral bounds for sparse {LDA}.
\newblock In \emph{Machine Learning, Proceedings of the Twenty-Third
  International Conference (ICML)}, pages 641--648, Pittsburgh, PA, 2006.

\bibitem[Roffo(2017)]{roffo2017ranking}
Giorgio Roffo.
\newblock Ranking to learn and learning to rank: On the role of ranking in
  pattern recognition applications.
\newblock \emph{arXiv:1706.05933}, 2017.

\bibitem[Roffo and Melzi(2016)]{roffo2016ranking}
Giorgio Roffo and Simone Melzi.
\newblock Ranking to learn.
\newblock In \emph{International Workshop on New Frontiers in Mining Complex
  Patterns}, pages 19--35. Springer, 2016.

\bibitem[Roffo et~al.(2015)Roffo, Melzi, and Cristani]{Proc:Roffo_ICCV15}
Giorgio Roffo, Simone Melzi, and Marco Cristani.
\newblock Infinite feature selection.
\newblock In \emph{2015 {IEEE} International Conference on Computer Vision
  (ICCV)}, pages 4202--4210, Santiago, Chile, 2015.

\bibitem[Roffo et~al.(2017)Roffo, Melzi, Castellani, and
  Vinciarelli]{Proc:Roffo_ICCV17}
Giorgio Roffo, Simone Melzi, Umberto Castellani, and Alessandro Vinciarelli.
\newblock Infinite latent feature selection: {A} probabilistic latent
  graph-based ranking approach.
\newblock In \emph{{IEEE} International Conference on Computer Vision (ICCV)},
  pages 1407--1415, Venice, Italy, 2017.

\bibitem[Safo et~al.(2018)Safo, Ahn, Jeon, and Jung]{safo2018sparse}
Sandra~E Safo, Jeongyoun Ahn, Yongho Jeon, and Sungkyu Jung.
\newblock Sparse generalized eigenvalue problem with application to canonical
  correlation analysis for integrative analysis of methylation and gene
  expression data.
\newblock \emph{Biometrics}, 2018.

\bibitem[Sleijpen and Van~der Vorst(2000)]{sleijpen2000jacobi}
Gerard~LG Sleijpen and Henk~A Van~der Vorst.
\newblock A {J}acobi--{D}avidson iteration method for linear eigenvalue
  problems.
\newblock \emph{SIAM Rev.}, 42\penalty0 (2):\penalty0 267--293, 2000.

\bibitem[Song et~al.(2015)Song, Babu, and Palomar]{song2015sparse}
Junxiao Song, Prabhu Babu, and Daniel~P Palomar.
\newblock Sparse generalized eigenvalue problem via smooth optimization.
\newblock \emph{IEEE Trans. Signal Process.}, 63\penalty0 (7):\penalty0
  1627--1642, 2015.

\bibitem[Sriperumbudur et~al.(2011)Sriperumbudur, Torres, and
  Lanckriet]{sriperumbudur2011majorization}
Bharath~K Sriperumbudur, David~A Torres, and Gert~RG Lanckriet.
\newblock A majorization-minimization approach to the sparse generalized
  eigenvalue problem.
\newblock \emph{Mach. Learn.}, 85\penalty0 (1-2):\penalty0 3--39, 2011.

\bibitem[Stewart(2001)]{stewart2001matrix}
Gilbert~W Stewart.
\newblock \emph{Matrix algorithms volume 2: eigensystems}, volume~2.
\newblock SIAM, 2001.

\bibitem[Tan et~al.(2018)Tan, Wang, Liu, and Zhang]{tan2018sparse}
Kean~Ming Tan, Zhaoran Wang, Han Liu, and Tong Zhang.
\newblock Sparse generalized eigenvalue problem: Optimal statistical rates via
  truncated {R}ayleigh flow.
\newblock \emph{J. R. Statist. Soc. B}, 80\penalty0 (5):\penalty0 1057--1086,
  2018.

\bibitem[Van~Loan and Golub(2012)]{van2012matrix}
Charles~F Van~Loan and Gene~H Golub.
\newblock \emph{Matrix Computations}.
\newblock Johns Hopkins University Press, Baltimore, MD, 4th edition, 2012.

\bibitem[Witten et~al.(2009)Witten, Tibshirani, and
  Hastie]{witten2009penalized}
Daniela~M Witten, Robert Tibshirani, and Trevor Hastie.
\newblock A penalized matrix decomposition, with applications to sparse
  principal components and canonical correlation analysis.
\newblock \emph{Biostatistics}, 10\penalty0 (3):\penalty0 515--534, 2009.

\bibitem[Yuan and Zhang(2013)]{yuan2013truncated}
Xiao-Tong Yuan and Tong Zhang.
\newblock Truncated power method for sparse eigenvalue problems.
\newblock \emph{J. Machine Learning Res.}, 14\penalty0 (Apr):\penalty0
  899--925, 2013.

\bibitem[Zou et~al.(2006)Zou, Hastie, and Tibshirani]{zou2006sparse}
Hui Zou, Trevor Hastie, and Robert Tibshirani.
\newblock Sparse principal component analysis.
\newblock \emph{J. Comput. Graphical Statist.}, 15\penalty0 (2):\penalty0
  265--286, 2006.

\end{thebibliography}

\onecolumn
\section*{Supplementary Materials}

\subsection*{A. Proof of Theorem 1}
\noindent{\bf Proof.}
Let $\lambda_{i,t+1}$ be the $i$th largest eigenvalue of $(\wt{A}_{\J_{t+1}},\wt{B}_{\J_{t+1}})$,
$\hat{\rho}^{(t+1)}$ be the same as in Lemma~3.
By the definition of $\eta_s^{(2)}$, we know that $\eta_s^{(2)}\ge \lambda_{2,t+1}$.
Together with $\rho^{(t)}>\eta_s^{(2)}$, we have $\rho^{(t)}>\lambda_{2,t+1}$.
On the other hand, using $|\J_t\cap\supp(v_1)|<k$, we know that $\rho^{(t)}\le\eta_{s,k-1}^{(1)}$.
Then by Lemma~3, we have
\begin{align*}
\lambda_{1,{t+1}}-\hat{\rho}^{(t+1)} \le ({\lambda}_{1,{t+1}} &- {\rho}^{(t)})\epsilon_m^2
+\mathcal{O}(({\lambda}_{1,t+1} - {\rho}^{(t)})^{\frac32}),
\end{align*}
where $\epsilon_m$ is the same as in Lemma~2.
By the definition of $\epsilon_*$, we know that $\epsilon_*\ge \epsilon_m$, it follows that
\begin{align*}
\lambda_{1,{t+1}}-\hat{\rho}^{(t+1)} \le ({\lambda}_{1,{t+1}} &- {\rho}^{(t)})\epsilon_*^2
+\mathcal{O}(({\lambda}_{1,t+1} - {\rho}^{(t)})^{\frac32}),
\end{align*}
Now using Lemma~4, we get the conclusion.
\qquad \hfill $\square$

\subsection*{B. Proof of Theorem 2}

\noindent{\bf Proof.}
Noticing that $|\supp(v^{(t)})|\le s$, using the definition of $\eta_{s,\ell}^{(1)}$, we know that
if $\rho^{(t)}>\eta_{s,k-1}^{(1)}$, then
\[
|\supp(v^{(t)})\cap\supp(v_1)|=k=|\supp(v_1)|.
\]
The conclusion follows immediately.
\qquad \hfill $\square$

\subsection*{C. Proof of Theorem 3}

In order to show Theorem~3, we need the following lemmas.



\begin{lemma}\label{lem:ghv}
Suppose $(A,B)$ is a symmetric-definite pair.
Let $E$, $F$ be two symmetric matrices with $\epsilon=\sqrt{\|E\|_2^2+\|F\|_2^2}<c(A,B)$.
Let $(\lambda,x)$ and $(\tilde{\lambda},\tilde{x})$ be the leading eigenpairs of $(A,B)$ and $(A+E,B+F)$, respectively.
Suppose $\tilde{\lambda}$ is simple,
and denote the smallest nonzero singular value of $(A+E)-\tilde{\lambda}(B+F)$ by $g$.
If $|\tilde{\lambda}|\epsilon < c(A,B)$,
then
\begin{align*}
\sin\theta(x,\tilde{x})\le \frac{\|B\|_2\delta+\sqrt{1+\tilde{\lambda}^2}\epsilon}{g},
\end{align*}
where
\begin{align}
\delta=\frac{(1+\tilde{\lambda}^2)\epsilon}{c(A,B)-|\tilde{\lambda}|\epsilon}.
\end{align}
\end{lemma}

\noindent{\bf Proof.}
First, since $\epsilon<c(A,B)$,
by Lemma~1, $(A+E,B+F)$ is a definite pair and 
\begin{align}\label{arctandiff}
\arctan(\tilde{\lambda}) -\arctan(\epsilon/c(A,B))\le  \arctan(\lambda)\le \arctan(\tilde{\lambda}) +\arctan(\epsilon/c(A,B)).
\end{align}
Using $|\tilde{\lambda}|\epsilon<c(A,B)$, we know that $\arctan(\epsilon/c(A,B))<\arctan(1/|\tilde{\lambda}|)=\frac{\pi}{2}-\arctan(|\tilde{\lambda}|)$,
which implies that the left hand side and righthand side of \eqref{arctandiff} are larger than $-\frac{\pi}{2}$ and smaller than $\frac{\pi}{2}$, respectively.
Then it follows from \eqref{arctandiff} that
\begin{align*}
\frac{\tilde{\lambda}c(A,B)-\epsilon}{c(A,B)+\tilde{\lambda}\epsilon}
\le\lambda\le
\frac{\tilde{\lambda}c(A,B)+\epsilon}{c(A,B)-\tilde{\lambda}\epsilon}.
\end{align*}
Therefore,
\begin{align}\label{lamdiff}
|\tilde{\lambda}-\lambda|\le \frac{(1+\tilde{\lambda}^2)\epsilon}{c(A,B)-|\tilde{\lambda}|\epsilon}=\delta.
\end{align}

Second,
without loss of generosity, we set $\|x\|_2=\|\tilde{x}\|_2=1$,
 let $r=[(A+E)-\tilde{\lambda}(B+F)]x$.
Direct calculations give rise to
\begin{align}
\|r\|_2&=\|(A-\tilde{\lambda}B)x+(E-\tilde{\lambda}F)x\|_2
\le \|(A-{\lambda}B)x\|_2+|\tilde{\lambda}-\lambda|\|Bx\|_2 +\|(E-\tilde{\lambda}F)x\|_2\notag\\
&\le \|B\|_2 \delta +\|E\|_2+|\tilde{\lambda}|\|F\|_2
\le \|B\|_2 \delta +\sqrt{1+\tilde{\lambda}^2}\epsilon.\label{rnorm}
\end{align}
On the other hand,
the spectral decomposition of $(A+E)-\tilde{\lambda}(B+F)$ can be given by
$(A+E)-\tilde{\lambda}(B+F)=V\diag(0,\gamma_2,\dots,\gamma_p)V^{\T}$,
where $V=[\tilde{x}, V_2]$ is orthogonal, $0>\gamma_2\ge\dots\ge\gamma_p$ are the eigenvalues of $(A+E)-\tilde{\lambda}(B+F)$.
Here we used the assumption that $\tilde{\lambda}$ is simple.
Then it follows that
\begin{align}\label{eq:v2ab}
V_2^{\T}r=V_2^{\T}[(A+E)-\tilde{\lambda}(B+F)]x=
\Gamma_2V_2^{\T}x,
\end{align}
where $\Gamma_2=\diag(\gamma_2,\dots,\gamma_p)$.
Using \eqref{rnorm} and \eqref{eq:v2ab}, we get
\begin{align*}
\sin\theta(x,\tilde{x})=\|V_2^{\T}x\|_2
=\|\Gamma_2^{-1}V_2^{\T}r\|_2
\le\frac{\|r\|_2}{|\gamma_2|}\le  \frac{\|B\|_2 \delta +\sqrt{1+\tilde{\lambda}^2}\epsilon}{g},
\end{align*}
which completes the proof.\qquad \hfill $\square$

\paragraph{Proof of Theorem 3.}\quad
Notice that  $(\lambda_1, (v_1)_{\J_t})$ and $(\rho^{(t)}, (v^{(t)})_{\J_t})$ are the leading eigenpairs of $(A_{\J_t},B_{\J_t})$ and $(\wt{A}_{\J_t},\wt{B}_{\J_t})$, respectively.
Then (a) and (b) follow from Lemma~1 and Lemma~\ref{lem:ghv}, respectively. This completes  the proof.
\qquad \hfill $\square$

%
\end{document}